\documentclass[letterpaper]{article} % DO NOT CHANGE THIS
\usepackage{aaai2026}  % DO NOT CHANGE THIS
\usepackage{times}  % DO NOT CHANGE THIS
\usepackage{helvet}  % DO NOT CHANGE THIS
\usepackage{courier}  % DO NOT CHANGE THIS
\usepackage[hyphens]{url}  % DO NOT CHANGE THIS
\usepackage{graphicx} % DO NOT CHANGE THIS
\usepackage{natbib}  % DO NOT CHANGE THIS AND DO NOT ADD ANY OPTIONS TO IT
\usepackage{caption} % DO NOT CHANGE THIS AND DO NOT ADD ANY OPTIONS TO IT
\usepackage{algorithm}
\usepackage{algorithmic}
\usepackage{enumitem}
\usepackage{amssymb}
\usepackage{mathtools}
\usepackage{amsthm}

\usepackage{amsmath}
\usepackage{transparent}
\usepackage{color}
\usepackage{xcolor}
\usepackage{colortbl}
\usepackage[makeroom]{cancel}
\usepackage{soul}  % background text color
\usepackage{pifont}% http://ctan.org/pkg/pifont
\usepackage{rotating} % rotatebox
\usepackage{float}
\usepackage{booktabs}

\definecolor{gray}{rgb}{0.46,0.46,0.46}
\definecolor{darkergreen}{RGB}{21, 152, 56}
\definecolor{darkerred}{RGB}{220, 35, 120}
\definecolor{darkerblue}{rgb}{0,0.08,0.45} % icml
\definecolor{royalblue}{RGB}{65,105,225}
\definecolor{lightblue}{RGB}{221,235,247}
\definecolor{gray94}{gray}{.94}
\definecolor{gray90}{gray}{.90}

\newcommand{\gray}[1]{\textcolor{gray}{#1}}

\newcolumntype{g}{>{\columncolor{gray94}}c} % column as gray
\newcolumntype{b}{>{\columncolor{lightblue}}c} % column as blue
\newcommand{\grow}[1]{\rowcolor{gray94}{#1}} % row as gray
\newcommand{\brow}[1]{\rowcolor{lightblue}{#1}} % row as blue
\newcommand{\bcell}[1]{\cellcolor{lightblue}{#1}} % cell as blue
\newcommand{\xmarkg}{\textcolor{gray}{\ding{55}}}%

\usepackage{newfloat}
\usepackage{listings}
\DeclareCaptionStyle{ruled}{labelfont=normalfont,labelsep=colon,strut=off} % DO NOT CHANGE THIS
\floatstyle{ruled}
\newfloat{listing}{tb}{lst}{}
\floatname{listing}{Listing}
\title{Life-Code: Central Dogma Modeling with Multi-Omics Sequence Unification}

\author{
    Zicheng Liu\textsuperscript{\rm 1,2,3}\thanks{Equal contribution.},
    Siyuan Li\textsuperscript{\rm 1,2,3}$^*$,
    Zhiyuan Chen\textsuperscript{\rm 4},
    Chang Yu\textsuperscript{\rm 1},\\
    Qirong Yang\textsuperscript{\rm 3},
    Yucheng Guo\textsuperscript{\rm 3},
    Yujie Yang\textsuperscript{\rm 3},
    Xiaoming Zhang\textsuperscript{\rm 3},
    Stan Z. Li\textsuperscript{\rm 1,3}\thanks{Corresponding author.}
}
\affiliations{
    \textsuperscript{\rm 1}AI Lab, Research Center for Industries of the Future, Westlake University, Hangzhou, China\\
    \textsuperscript{\rm 2}Zhejiang University, Hangzhou, China\\
    \textsuperscript{\rm 3}BioMap Research, Beijing, China\\
    \textsuperscript{\rm 4}University of Hong Kong, Hong Kong, China\\

}

\usepackage{bibentry}
\begin{document}

\maketitle

%========= BODY TEXT =========
\begin{abstract}

The interactions between DNA, RNA, and proteins are fundamental to biological processes, as illustrated by the central dogma of molecular biology.
Although modern biological pre-trained models have achieved great success in analyzing these macromolecules individually, their interconnected nature remains underexplored.
This paper follows the guidance of the central dogma to redesign both the data and model pipeline and offers a comprehensive framework, Life-Code, that spans different biological functions. As for data flow, we propose a unified pipeline to integrate multi-omics data by reverse-transcribing RNA and reverse-translating amino acids into nucleotide-based sequences. As for the model, we design a codon tokenizer and a hybrid long-sequence architecture to encode the interactions between coding and non-coding regions through masked modeling pre-training.
To model the translation and folding process with coding sequences, Life-Code learns protein structures of the corresponding amino acids by knowledge distillation from off-the-shelf protein language models.
Such designs enable Life-Code to capture complex interactions within genetic sequences, providing a more comprehensive understanding of multi-omics with the central dogma.
Extensive experiments show that Life-Code achieves state-of-the-art results on various tasks across three omics, highlighting its potential for advancing multi-omics analysis and interpretation.

\end{abstract}
\section{Introduction}
\begin{figure}[t]
    \centering
    \vspace{-0.5em}
    \includegraphics[width=1.0\linewidth]{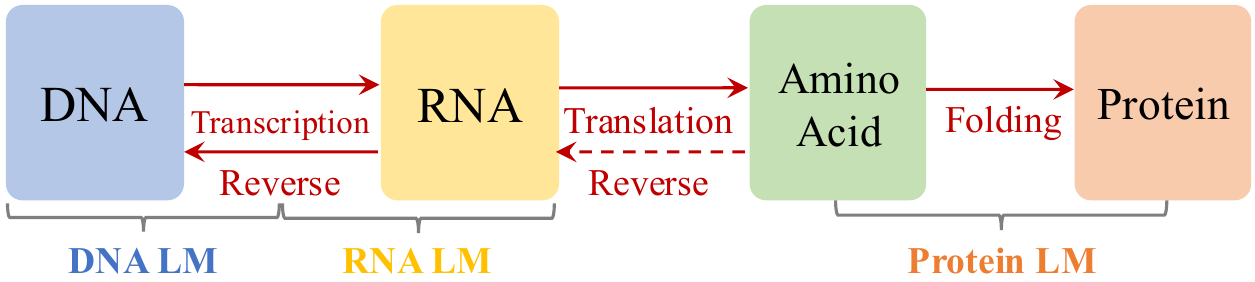}
    \vspace{-1.75em}
    \caption{\textbf{Overview of the Central Dogma Data Flow} with biological language models (LM), separately proposed.
    }
    \label{fig:central_dogma}
    \vspace{-1.0em}
\end{figure}

The advent of large-scale biological datasets has sparked extensive interest in \textit{sequence modeling} across multiple omics domains---ranging from \textit{DNA} and \textit{RNA} to \textit{proteins}~\cite{liwang2024protein,icml2024vqdna,liu2024genbench}.
These endeavors aim to leverage deep learning to uncover functional elements, predict gene regulation, and accelerate protein engineering. Notably, the successful application of \textit{Transformer}-based architectures in natural language processing (NLP)~\cite{vaswani2017attention, devlin2019bert} has inspired numerous efforts to adapt these methods to \textit{biology}, giving rise to models for variant effect prediction~\cite{zhou2022transformer}, protein structure elucidation~\cite{jumper2021alphafold}, and more.
Despite the promise of these paradigms of foundation models in computational biology~\cite{lin2023esm,nguyen2024evo,shen2024rnafm}, three major obstacles continue to hamper \textit{multi-omics} modeling:
\textbf{(1) Data Island.} Current approaches often focus on \textit{one molecular modality}, modeling only DNA, RNA, or only protein---leading to siloed representations that cannot exploit the inherent relationships among DNA, RNA, and proteins~\cite{hong2022crossomics}, as shown in Figure~\ref{fig:central_dogma}. As a result, critical cross-omics information (for example, how a genetic variant in DNA might alter RNA splicing and ultimately impact protein function) remains underutilized.
\textbf{(2) Na\"ive ``Copy and Paste''.} While Transformers~\cite{vaswani2017attention, devlin2019bert} have revolutionized NLP, simply applying them without incorporating \textit{biological knowledge} can undermine performance and interpretability~\cite{rives2021biological}. For instance, ignoring codons, coding, and non-coding regions may limit model accuracy and obscure how predictions map to real biological processes.
\textbf{(3) Inefficiency for Long Sequences.} Genomic and transcriptomic sequences can span \textit{thousands to millions} of bases, and mapping proteins back to nucleotide triplets multiplies sequence length~\cite{alquraishi2019proteinnet}. This expansion quickly renders the \(\mathcal{O}(n^2)\) complexity of vanilla Transformers computationally prohibitive, especially for large-scale training on full genomes.

\begin{figure*}[t!]
    \vspace{-0.25em}
    \centering
    \includegraphics[width=0.9\linewidth]{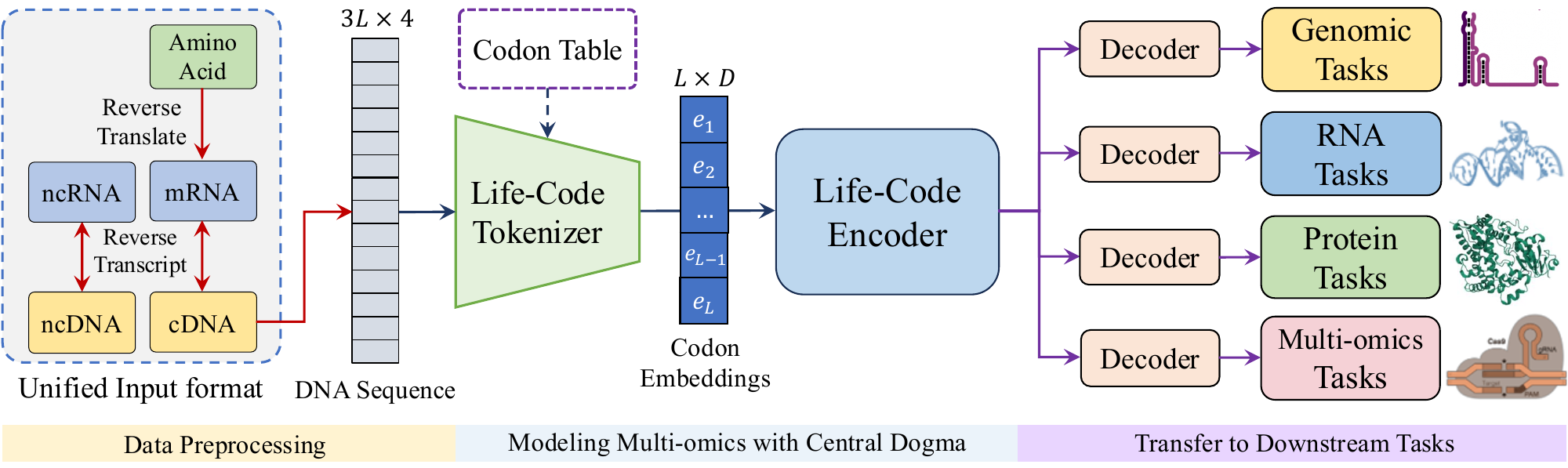}
    \vspace{-0.5em}
    \caption{\textbf{Illustration of Life-Code Framework}, which contains three pipelines. (a) \textbf{Data Pre-processing} for the unified input with the central dogma. (b) \textbf{Pre-training} of the Tokenizer and the hybrid Encoder to model the contextual information and the central dogma rules. (c) \textbf{Transfer Learning} to multi-omics downstream tasks with parameter-efficient Supervised Fine-tuning.
    %%%%%%% color version %%%%%%%
    % \caption{\textbf{Illustration of Life-Code Framework}, which contains three pipelines. (a) \yellow{\textbf{Data Pre-processing}} for the unified input with the central dogma. (b) \blue{\textbf{Pre-training}} of the Tokenizer and the hybrid Encoder to model the contextual information and the central dogma rules. (c) \purple{\textbf{Transfer Learning}} to multi-omics downstream tasks with parameter-efficient Supervised Fine-tuning.
    }
    \label{fig:lcode_framework}
    \vspace{-1.0em}
\end{figure*}

% \begin{wrapfigure}{r}{0.5\textwidth}
%     \vspace{-0.25em}
%     \centering
%     \includegraphics[width=1.0\linewidth]{figs/fig_intro.pdf}
%     % \vspace{-2.25em}
%     \caption{\textbf{Overview of the Central Dogma Data Flow} with biological language models (LM) severally proposed in each omics.
%     }
%     \label{fig:central_dogma}
%     \vspace{-1.0em}
% \end{wrapfigure}
% \input{tabs/tab_overall}
This paper introduces \textbf{Life-Code}, as shown in Figure~\ref{fig:lcode_framework}, a unifying framework that addresses these limitations through \textit{multi-omics unification}, \textit{bio-interpretable designs}, and \textit{long-sequence efficiency}. By grounding our method in the \textit{central dogma} of molecular biology, we align DNA, RNA, and protein sequences under a single representation that captures cross-omics signals. We then augment this representation with a \textit{hybrid} backbone architecture, where key biological structures (\textit{e.g.}, coding sequences, promoters, and regulatory motifs) guide the model to focus computational resources effectively. In particular, we explore specialized attention mechanisms that mitigate the memory bottleneck of na\"ive Transformer models while retaining sufficient global context for \textit{biologically relevant} long-range dependencies.
In summary, this work makes four \textbf{key contributions} to the field of computational biology with deep learning:

% \begin{itemize}
\begin{itemize}[leftmargin=2.0em]
    \vspace{-0.5em}
    \item \textbf{Multi-omics Modeling.} 
    We demonstrate a novel \textit{multi-omics} approach by co-representing DNA, RNA, and protein within a \textit{single} model, thus overcoming the traditional data island problem and enabling the transfer of knowledge across modalities.
    
    \vspace{-2pt}
    \item \textbf{Bio-interpretable Design.} 
    Our tokenization and training pipeline explicitly distinguishes \textit{coding (CDS)} and \textit{non-coding (nCDS)} regions, preserving \textit{biological interpretability}. Through tasks like masked language modeling for nCDS and protein translation objectives for CDS, the model learns to capture both regulatory and translational signals.

    \vspace{-2pt}
    \item \textbf{Efficient Long-sequence Computing.}
    We combine structured prior knowledge (\textit{e.g.}, double-helix complementarity, coding frames) with efficient linear attention mechanisms---balancing representational power with reduced complexity for long sequences. This hybrid strategy yields significant computational savings over na\"ive \(\mathcal{O}(n^2)\) Transformer approaches.

    \vspace{-2pt}
    \item \textbf{Comprehensive Validations.}
    Evaluated on diverse tasks in DNA, RNA, and protein applications, empirical results confirm the generalizability of Life-Code in these fundamental modalities.
\end{itemize}

\section{Method}
\subsection{Preliminaries}

\paragraph{Central Dogma links islands.}
The \textit{central dogma} of molecular biology describes the flow of genetic information from DNA $\rightarrow$ RNA $\rightarrow$ Protein. DNA is transcribed into RNA, and RNA is subsequently translated into protein. This process ensures that the information encoded in the genome can be expressed to perform diverse biological functions. Let \( \mathcal{D} \subseteq \{\texttt{A}, \texttt{T}, \texttt{C}, \texttt{G}\}^*\) be the space of DNA sequences, and let \(\mathcal{R} \subseteq \{\texttt{A}, \texttt{U}, \texttt{C}, \texttt{G}\}^*\) be the space of RNA sequences. For proteins, denote \(\mathcal{P} \subseteq \Sigma^*\), where \(\Sigma\) is the amino acid alphabet (\textit{e.g.}, \(\Sigma = \{ \texttt{Ala}, \texttt{Arg}, \texttt{Asn}, ...\}\)). The central dogma stipulates two fundamental mappings:
\begin{align}
    T_{\mathrm{transcribe}}: \mathcal{D} \to \mathcal{R}, \quad
    T_{\mathrm{translate}}: \mathcal{R} \to \mathcal{P}.
\end{align}
By composition, one obtains \(T_{\mathrm{translate}} \circ T_{\mathrm{transcribe}}: \mathcal{D} \to \mathcal{P}\). In practice, DNA $\rightarrow$ RNA $\rightarrow$ Protein is not a strict one-to-one mapping due to splicing, alternative start sites, etc. However, these transformations offer critical biological correspondences that can link data from different omics sources.
\textit{Why it matters}: By grounding sequences in DNA (\( \mathbf{x} \in \mathcal{D} \)), we naturally align each transcript (\( \mathbf{y} \in \mathcal{R} \)) and its resultant protein (\( \mathbf{z} \in \mathcal{P} \)) within a unified coordinate system. This approach not only enlarges the effective dataset via cross-modality but also preserves mechanistic interpretability, as variations in \(\mathbf{x}\) manifest as changes in \(\mathbf{y}\) and \(\mathbf{z}\).

\paragraph{The bi-directional hybrid model is more efficient.}
DNA has three notable properties:
(1) Length: A chromosome-scale DNA sequence \(\mathbf{x} = (x_1, x_2, \ldots, x_n)\) can reach millions (or more) of base pairs (bp).
(2) Double-Helix Complementarity: Each position \(x_i \in \{\texttt{A}, \texttt{T}, \texttt{C}, \texttt{G}\}\) pairs with a complementary base on the opposite strand.
(3) Distributed Regulatory Elements: Important functional sites (\textit{e.g.}, promoters, enhancers, splicing signals) can be located in different regions (CDS or nCDS). In summary,  while Transformer-based architectures exhibit strong representational power, their attention mechanism scales with \(\mathcal{O}(n^2)\) complexity, making it difficult to handle sequences of length \(n\gg 10^4\). Linear or sparse approximations (\textit{e.g.}, Mamba) reduce complexity but often sacrifice representational fidelity for complex biological contexts. Hence, the recent efficient hybrid strategy~\cite{yang2025deltanet,yang2024gateddelta, iclr2024MogaNet,icml2024chela,ijcai2024longvq} should be considered.
We now formally describe how we unify DNA, RNA, and protein sources into \textit{nucleotides}.

\subsection{Data Pipeline for Life-Code}
\paragraph{Unified to nucleotides.} Given an RNA sequence \(\mathbf{y} \in \mathcal{R}\) or a protein sequence \(\mathbf{z} \in \mathcal{P}\), we map them back to DNA-level tokens. Specifically: For RNA, assume we have the corresponding genomic DNA region (\textit{e.g.}, from reference genome or aligned reads) and apply \(\mathbf{y} = T_{\mathrm{transcribe}}(\mathbf{x})\) in reverse to identify the original \(\mathbf{x} \in \mathcal{D}\). For protein, we leverage the standard codon translation table. Each amino acid \(z_j \in \Sigma\) typically maps to one or more codons in \(\{\texttt{A}, \texttt{T}, \texttt{C}, \texttt{G}\}^3\). We choose the canonical codon (or the most frequent codon) for each amino acid, yielding a surjective function
    \(g: \Sigma \to \{\texttt{A}, \texttt{T}, \texttt{C}, \texttt{G}\}^3.\)
Thus, a protein of length \(L\) becomes a DNA-like sequence of length \(3L\).

\begin{figure}[t!]
    % \vspace{-1em}
    \centering
    \includegraphics[width=1.0\linewidth]{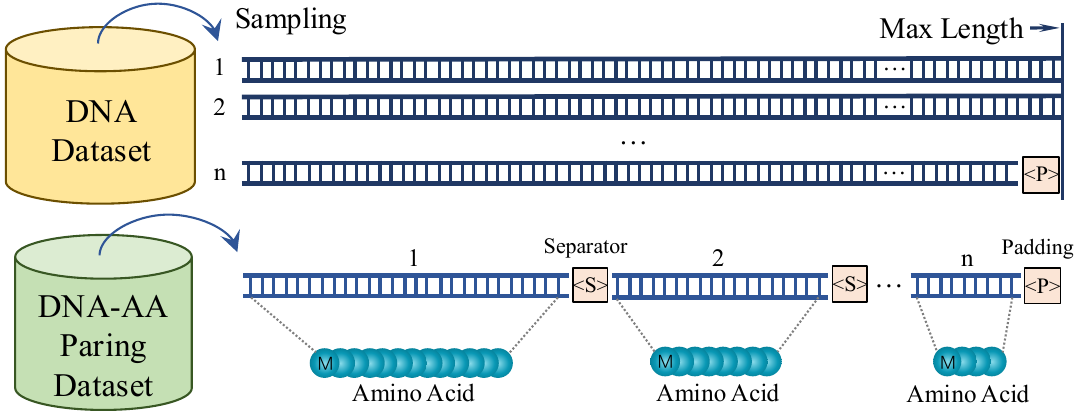}
    % \vspace{-2.0em}
    \caption{\textbf{Data Sampling Pipeline}. During pre-training, we sample the DNA sequences and the pair of CDS and amino acids from the DNA dataset and the DNA-AA pairing dataset. Since the DNA sequence sampled from the RefSeq is much longer than the CDS, we predefined the max sequence length (\textit{e.g.}, 8k) for each sample. As for the CDS and its corresponding amino acids, we apply the packing strategy~\cite{Warner2024ModernBERT} to sample with several CDSs.
    }
    \label{fig:data_packing}
    \vspace{-1em}
\end{figure}
\paragraph{Data sampling.}
Following GenBank~\cite{benson2012genbank} curation, we sample from the \texttt{RefSeq} database, truncating long genomic segments and padding shorter ones. For \textit{CDS--Amino Acid} pairs, we extract annotated CDS records and their corresponding protein sequences, then pack multiple short segments into a single training sample with special separators. See Figure~\ref{fig:data_packing} for a schematic illustration.
This approach enables us to efficiently process both longer genomic segments and multiple shorter CDS-AA pairs in a unified framework. By balancing \emph{truncation/padding} for DNA sequences and \emph{packing} for CDS segments, we ensure that each batch contains a diverse mixture of long-range genomic context and coding information, thereby maximizing the model’s exposure to various biological signals.

% \subsection{Tokenizer Training.} 
% After mapping proteins and RNA to DNA-like sequences, all data can be treated as elements of \(\{\texttt{A}, \texttt{T}, \texttt{C}, \texttt{G}\}^*\). To reduce the length of the protein-mapped sequences (which can balloon by a factor of 3), we apply a stride-3 convolution to compress each triplet:
% \[
% \mathbf{u}_j = \mathrm{Conv}(\mathbf{x}_{3j+1:3j+3}),
% \]
% where \(\mathbf{u}_j\) is a compressed token embedding. Concretely, if the mapped protein segment has length \(3L\), this yields an embedding sequence of length \(L\). Finally, we pass each token embedding \(\mathbf{u}_j\) through a small MLP:
% \[
% \mathbf{e}_j = \mathrm{MLP}(\mathbf{u}_j),
% \]
% resulting in the final token embedding \(\mathbf{e}_j \in \mathbb{R}^d\).  
% Thus, all data—DNA, RNA, or protein—can be \textit{unified} into a single nucleotide-level embedding space.

\subsection{Life-Code Tokenizer Training}

\begin{figure}[t!]
    \centering
    \includegraphics[width=0.9\linewidth]{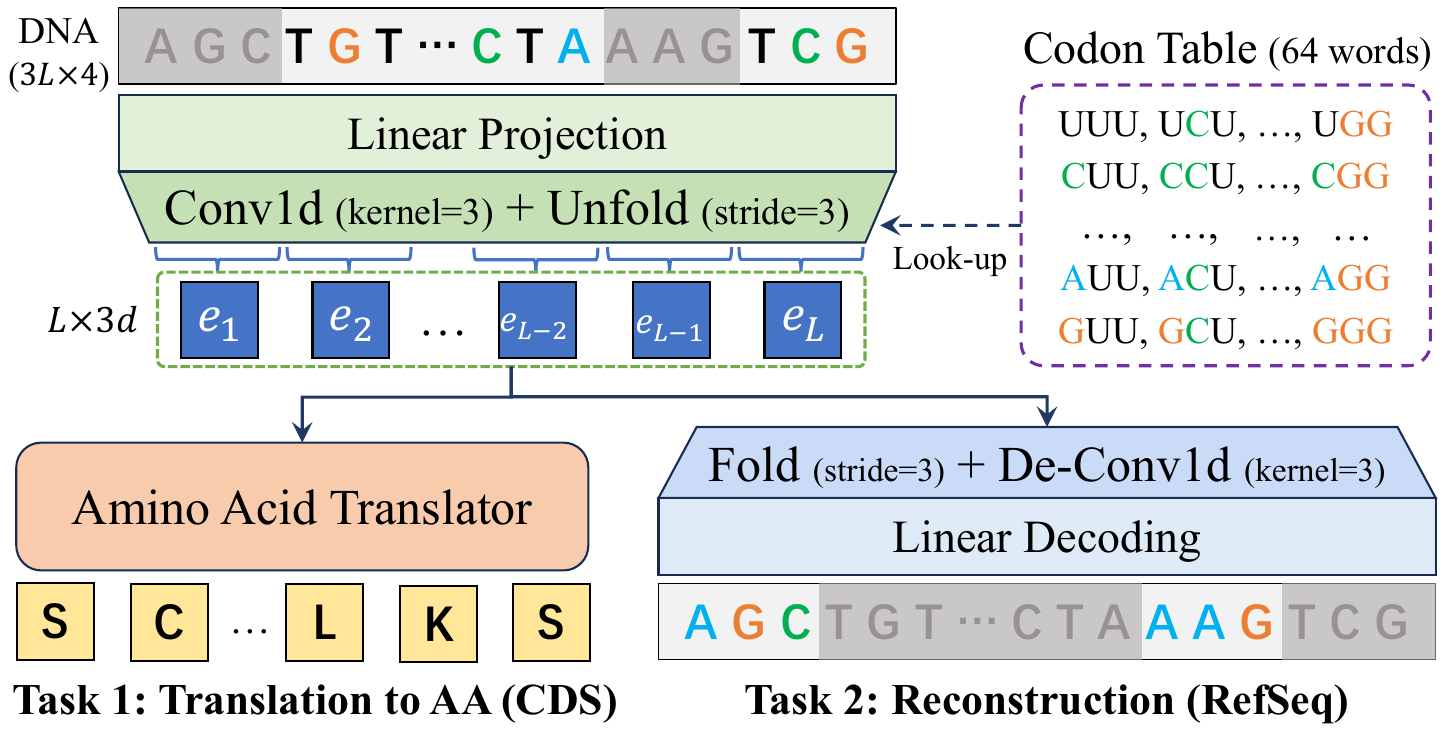}
    \vspace{-0.5em}
    \caption{\textbf{Life-Code Tokenizer}. To model the interactions among DNA, RNA, and Amino Acids, our tokenizer takes nucleotide acids (4 words) as the inputs, then takes codons as the latent vocabulary (64 words), and translates them to Amino Acids (20 words) as the outputs, which could be pre-trained by (a) nucleotide masked modeling and (b) CDS to Amino Acid translation.
    }
    \label{fig:tokenizer}
    \vspace{-1em}
\end{figure}

After mapping all sequences (DNA, RNA, and protein) to a unified DNA-like form, each sample can be viewed as a string in \(\{\texttt{A}, \texttt{T}, \texttt{C}, \texttt{G}\}^*\). However, to effectively capture both nucleotide-level and codon-level patterns, as shown in Figure~\ref{fig:tokenizer}, our tokenizer incorporates a \textit{two-task} training paradigm: (\emph{i}) masked language modeling for nucleotide reconstruction and (\emph{ii}) protein translation for CDS.

\paragraph{Tokenization with codon-level embeddings.}
Consider a raw DNA segment of length \(3L\). We first apply a 1-d convolution (kernel size \(=3\)) together with an ``unfold'' operation (stride \(=3\)) to transform each consecutive triplet \((x_{3j+1}, x_{3j+2}, x_{3j+3})\) into a codon-like representation:
\[
\mathbf{e}_j \;=\; \mathrm{Conv1d}(\,\mathbf{x}_{3j+1:3j+3}\,).
\]
This step effectively condenses triplets---\emph{codons}---into a more compact embedding sequence \(\{\mathbf{e}_1, \mathbf{e}_2, \dots, \mathbf{e}_L\} \in \mathbb{R}^{L\times 3d}\). We apply a small linear projection or MLP to each \(\mathbf{e}_j \in \mathbb{R}^{d}\), mapping it into the token embedding space.

\paragraph{Dual pre-training.}
\label{sec:tokenizer_tasks}
We jointly pre-train this tokenizer on two objectives:
(1) Nucleotide MLM (Reconstruction).
    For each codon-embedded sequence, we randomly mask a subset of tokens and train the model to reconstruct the original nucleotides. Technically, we design a \textit{DNA De-Tokenizer} that applies a \texttt{Fold} (stride = 3) and an inverse convolution (\texttt{De-Conv1d}) to recover the three-base codon, thereby reconstructing the full-length of DNA sequences. Formally, if \(\widehat{\mathbf{x}}_{3j+1:3j+3}\) is the predicted codon for position \(j\), the loss is:
    \begin{equation}
        \label{eq:loss_MLM}
        \mathcal{L}_{\mathrm{MLM}} \;=\; 
        \sum_{j=1}^L 
        -\log p\bigl(\mathbf{x}_{3j+1:3j+3} \,\big\vert\, \widetilde{\mathbf{x}}\bigr),
    \end{equation}
    where \(\widetilde{\mathbf{x}}\) is the masked sequence. This ensures the tokenizer learns robust nucleotide representations that can reconstruct the original sequence.
(2) CDS-to-Protein Translation.
    For coding regions, each codon corresponds to an amino acid (\(\mathbf{z}_j \in \Sigma\)). We thus include the \textit{Amino Acid Translator} module, a two-layer MLP layer, that looks up the genetic code table and learns to predict \(\mathbf{z}_j\) given \(\mathbf{e}_j\):
    \begin{equation}
        \label{eq:loss_trans}
        \mathcal{L}_{\mathrm{Trans}} \;=\; 
        \sum_{j=1}^L 
        -\log\,p\bigl(\mathbf{z}_j \,\big\vert\, \mathbf{e}_j\bigr).
    \end{equation}
    By training on paired CDS--amino-acid data, the tokenizer acquires biologically grounded representations, effectively bridging codon-level features to protein-level semantics.

\subsection{Bi-Directional Hybrid Model for Long Sequence}
\label{sec:hybrid-model}

We describe the \textbf{Life-Code Encoder}, which takes the tokenized embeddings as input and models DNA sequences efficiently, respecting double-strand complementarity.

\paragraph{Bi-directional input.}
To use the double-helix property as Caduceus~\citep{icml2024caduceus}, we split the embeddings \(\mathbf{E} = (\mathbf{e}_1, \mathbf{e}_2, \ldots, \mathbf{e}_n)\) into two parts in the \textit{feature} dimension:
\[
(\mathbf{E}^{(+)},\ \mathbf{E}^{(-)}) = \mathrm{Split}(\mathbf{E}),
\]
where \(\mathbf{E}^{(+)}\) represents the forward strand embedding, and \(\mathbf{E}^{(-)}\) represents the reverse (complementary) strand. We process these two strands in parallel using the same model weights or parameter-sharing scheme \(f_\theta\):
\begin{align*}
    \mathbf{H}^{(+)} = f_\theta(\mathbf{E}^{(+)}), \quad
    \mathbf{H}^{(-)} = f_\theta(\mathrm{Reverse}(\mathbf{E}^{(-)})).
\end{align*}
Finally, we concatenate or fuse the representations:
\[
\mathbf{H} = [\mathbf{H}^{(+)},\ \mathbf{H}^{(-)}],
\]
where $\mathbf{H} \in \mathbb{R}^{D}$ encodes both the forward and reverse complement features.

\begin{figure*}[t!]
    \centering
    \includegraphics[width=1\linewidth]{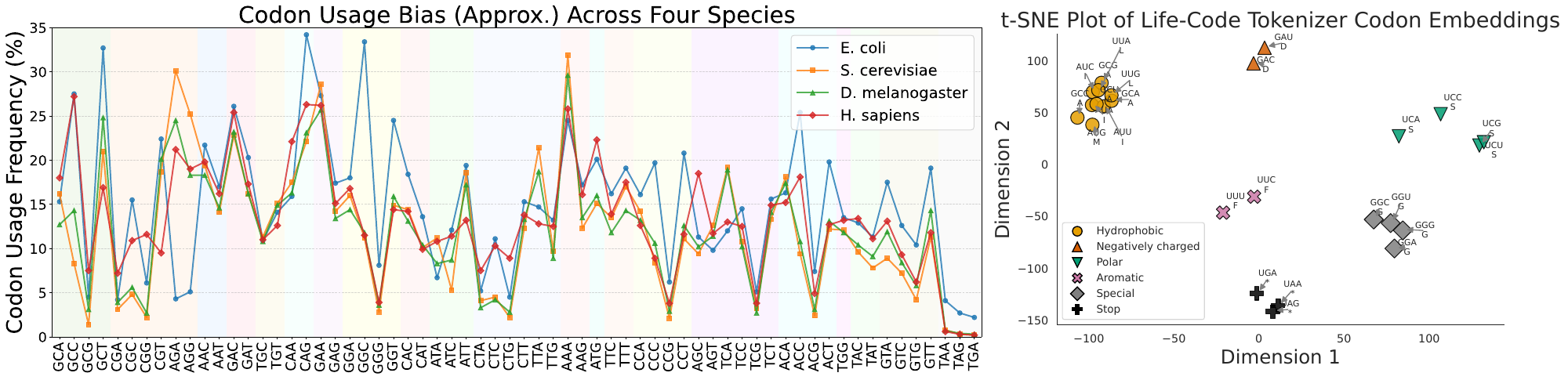}
    \vspace{-1.5em}
    \caption{\textbf{Empirical Analysis of Life-Code Tokenizer}.
    \textbf{Left:} \textit{Codon Usage Bias} in Life-Code Tokenizer across four representative species---\textit{E.~coli}, \textit{S.~cerevisiae} (yeast), \textit{D.~melanogaster} (fruit fly), and \textit{H.~sapiens} (human)---illustrating variations in codon frequency (\%) for amino acids. The pastel strips highlight codons belonging to the same amino acid group. 
    \textbf{Right:} \textit{t-SNE visualization} of learned codon embeddings in Life-Code Tokenizer, where codons that translate to the same or biochemically similar amino acids cluster together (\textit{e.g.}, hydrophobic or charged groups), and stop codons form a distinct region.
    }
    \vspace{-1.5em}
    \label{fig:codon}
\end{figure*}

\paragraph{Hybrid Token Mixer.}
Inspired by recent advances in EVO-style architectures~\cite{nguyen2024evo} and Gated Delta Networks~(GDN)~\cite{yang2024gateddelta}, we design the encoder in Life-Code as the linear-attention Gated DeltaNet \emph{mixed} with a small proportion of standard multi-head self-attention (MHSA). Concretely, for every twelve encoder layers, eleven employ the GDN update rule---offering linear or near-linear complexity via kernel-based reordering---while one layer applies MHSA. The GDN layer relies on a \emph{delta update} to maintain memory, reducing the quadratic overhead of naive Transformers. Meanwhile, the \emph{one} full-attention layer provides additional global context, which is crucial for capturing fine-grained long-range dependencies. This \textit{11\,{:}\,1} ratio of GDN to dense layers significantly boosts computational efficiency (compared to a pure \(\mathcal{O}(n^2)\) Transformer) while preserving model expressiveness and long-distance modeling capacity.

\begin{figure*}[t!]
    \vspace{-1em}
    \centering
    \includegraphics[width=1.0\linewidth]{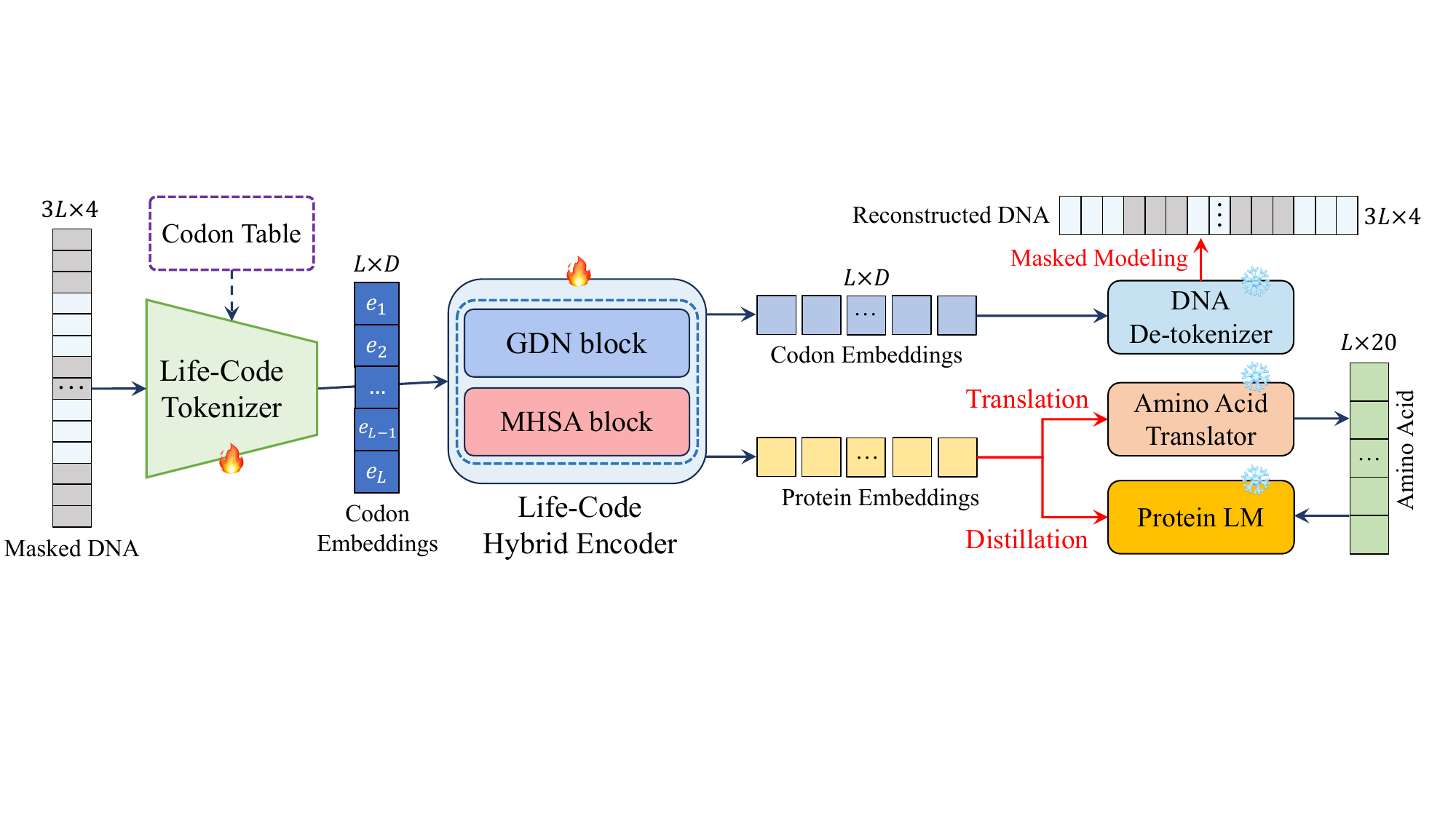}
    \vspace{-1.75em}
    \caption{\textbf{Illustration of Life-Code Encoder pre-training pipeline}. We model the long sequence by the hybrid encoder as a mixture of DeltaNet and Attention blocks. With pre-trained tokenizers and de-tokenizers, we consider three types of pre-training tasks: (a) the nucleotide masked modeling for contextual information, (b) the cDNA-AA translation for the central dogma, and (c) the knowledge distillation (KD) with pre-trained protein LM.
    Note that red lines indicates the learning objectives.
    %%%%%%% color version %%%%%%%
    % Note that \red{red} lines indicates the learning objectives.
    }
    \label{fig:lcode_pretraining}
    \vspace{-1.0em}
\end{figure*}

\paragraph{Encoder Pre-training.}
Once the decoder and translator of our tokenizer weights are fixed (see Section~\ref{sec:tokenizer_tasks}), we train the \emph{Life-Code Encoder} (Figure~\ref{fig:lcode_pretraining}) on three tasks: (1) Masked DNA Reconstruction (MLM), (2) CDS-to-Amino-Acid Translation, and (3) Token-Level Knowledge Distillation (KD) from a pretrained Protein LM. The first two objectives match those used to pre-train the tokenizer but are now applied to the encoder output. Specifically, the encoder generates codon embeddings that a \emph{DNA De-tokenizer} can reconstruct (MLM) and that an \emph{Amino Acid Translator} can convert to protein residues (Translation). 

For the \textbf{distillation} step, we replace the previous squared-error formulation with a \emph{token-wise KL loss} on the predicted distributions, which we have empirically found to be more robust (\textit{e.g.}, no negative log issues). Inspired by the DINO framework~\citep{oquab2024dinov2}, we \emph{decode} the encoder’s final protein sequence embeddings to protein structure embeddings a \textit{Protein Decoder} $\mathcal{D}$ to produce a per-token distribution aligned with the teacher PLM like ESM2~\citep{lin2022ESM2}. Formally, let \(P_\theta(\mathcal{D} (\mathbf{h}_j^\mathrm{LifeCode})) \in \mathbb{R}^{D'}\) and \(P_{\theta'}(\mathbf{h}_j^\mathrm{PLM}) \in \mathbb{R}^{D'}\) be the predicted distributions for token \(j\) from our model and the teacher, where the \textit{Decoder} can be a two-layer-MLP with \(D \rightarrow 4D \rightarrow D'\). Then, the \emph{alignment} or KD loss is defined as:
\begin{equation}
    \label{eq:tokenwise_kl}
    \mathcal{L}_\mathrm{KD} \;=\; \sum_{j=1}^{L} 
    -P_{\theta'}\bigl(\mathbf{h}_j^\mathrm{PLM}\bigr)^\top \log P_{\theta}\bigl(\mathcal{D}( \mathbf{h}_j^\mathrm{LifeCode}) \bigr).
\end{equation}
We combine these three losses using a \textit{log-sum} multi-task strategy~\citep{lin2023dualmtl} so that all terms (MLM, Translation, and KD) are treated as entropy-like objectives:
\begin{equation}
    \hspace{-1.0em}
    \label{eq:loss_total}
    \log\bigl(\mathcal{L}_\mathrm{MLM} + \epsilon\bigr) 
    \;+\; \log\bigl(\mathcal{L}_\mathrm{Trans} + \epsilon\bigr)
    \;+\; \log\bigl(\mathcal{L}_\mathrm{KD} + \epsilon\bigr),
\end{equation}
where \(\epsilon\) is a small constant. This adaptive weighting prevents any single task from dominating and ensures stable optimization across various training stages. Notably, we also adopt this strategy when training the tokenizer.

\begin{figure}[b!]
    \vspace{-1em}
    \centering
    \includegraphics[width=0.9\linewidth]{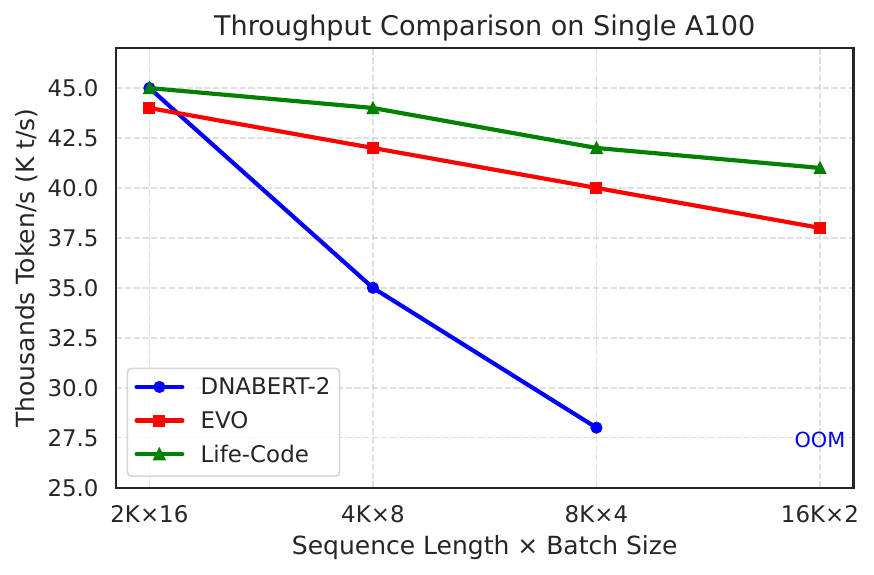}
    \vspace{-1em}
    \caption{%
    \textbf{Training Efficiency on a Single A100 GPU.}
    We measure throughput (K t/s) at four sequence lengths–batch size settings.
    DNABERT-2 slows significantly at higher lengths and reaches an out-of-memory (OOM) at \(16\text{K}\times2\).
    }
    \label{fig:efficiency}
    \vspace{-0.5em}
\end{figure}

\subsection{Empirical Consequences}
\paragraph{Biological insights}
In Figure~\ref{fig:codon}, our analyses reveal that the Life-Code Tokenizer captures biologically meaningful codon relationships in two complementary ways. First, codon usage bias (left figure) indicates that different species, from \emph{E. coli} to \emph{H. sapiens}, vary markedly in their preference for certain codons—an effect traditionally tied to tRNA abundance and translational efficiency. By visualizing these preferences across amino acids, we observe how each organism clusters around specific high-frequency codons, underscoring the ecological and evolutionary factors shaping codon selection.
Second, the t-SNE plot of the tokenizer’s learned codon embeddings (right figure) demonstrates that codons translating to amino acids with similar biochemical properties (\textit{e.g.}, hydrophobic or charged) are embedded close together, while the stop codons occupy a separate region. This structural arrangement confirms that the model internalizes codon–amino acid mappings, reflecting both frequency patterns and higher-level attributes like polarity or aromaticity. Together, these findings suggest that the Life-Code Tokenizer not only memorizes sequence data but also discerns and encodes intrinsic biological signals.

%%%%%%% color version %%%%%%%
% \paragraph{Training efficiency}
% We evaluated the throughput (thousands of tokens processed per second) of Life-Code, EVO, and DNABERT-2 on a single A100 GPU under increasing sequence lengths (with batch size adjusted accordingly). As shown in Figure~\ref{fig:efficiency}, DNABERT-2 (\blue{blue}) exhibits a steep drop-off at longer sequences and ultimately runs out of memory (OOM) at \(16\text{K}\times2\). In contrast, EVO (\red{red}) and Life-Code (\green{green}) both maintain higher throughput thanks to their hybrid attention mechanisms. Notably, Life-Code achieves the best scalability, retaining above 40K t/s even at \(8\text{K}\times4\). These results confirm that Life-Code’s efficient architecture can accommodate extensive input lengths with minimal degradation in training speed.
\paragraph{Training efficiency}
We evaluated the throughput (thousands of tokens processed per second) of Life-Code, EVO, and DNABERT-2 on a single A100 GPU under increasing sequence lengths (with batch size adjusted accordingly). As shown in Figure~\ref{fig:efficiency}, DNABERT-2 (blue) exhibits a steep drop-off at longer sequences and ultimately runs out of memory (OOM) at \(16\text{K}\times2\). In contrast, EVO (red) and Life-Code (green) both maintain higher throughput thanks to their hybrid attention mechanisms. Notably, Life-Code achieves the best scalability, retaining above 40K t/s even at \(8\text{K}\times4\). These results confirm that Life-Code’s efficient architecture can accommodate extensive input lengths with minimal degradation in training speed.

\section{Experiments}
\label{sec:exp}
We first introduce the network architectures and pre-training settings of our Life-Code tokenizer and encoder. Then, we evaluate Life-Code on DNA, RNA, protein, and multi-omics tasks with supervised fine-tuning (SFT) or zero-shot evaluation protocols. All experiments are conducted with PyTorch, \textit{transformers} library, and NVIDIA A100-80G GPUs. 

\subsection{Experimental Setup}
\label{sec:exp_setting}
% \vspace{-0.25em}
\paragraph{Life-Code Tokenizer.}
With the nucleotide vocabulary, the \textit{Life-Code tokenizer} contains the following modules: (\romannumeral1) A linear projection from the DNA input to 384-dim implemented by \texttt{nn.Embedding}; (\romannumeral2) A GDN block (DeltaNet) with 384-dim for global contextual modeling; (\romannumeral3) A 1-d Convolution with a kernel size of 3 and stride of 1, followed by an \texttt{UnFold} operation to merge every three nucleotide tokens into 768-dim codon embedding. Similarly, we also design the \textit{DNA De-Tokenizer} and \textit{Amino Acid Translator} for tokenizer pre-training.
As shown in Figure~\ref{fig:tokenizer}, we first pre-train the Life-Code tokenizer by AdamW optimizer~\citep{iclr2019AdamW} for 100,000 iterations with a basic learning rate of $2\times 10^{-4}$ adjusted by the Cosine scheduler and a total batch size of 512, as summarized in Table~\ref{tab:app_lcode_config}. As shown in Table~\ref{tab:app_dataset} and ~\ref{fig:data_collection}, the two pre-training datasets (DNA and DNA-AA pairing) are constructed from NCBI, GenBank, and UniRef50 databases.
% View Appendix~\ref{app:impl_tokenizer} for details.

\vspace{-0.25em}
\paragraph{Life-Code Encoder.}
Similar to the BERT-Large architecture~\citep{devlin2019bert}, the Life-Code encoder has 24 layers in total with an embedding dim of 1024/1280 and 340/500M parameters. We enhance the model with three-fold designs: (\romannumeral1) Mixture of DeltaNet~\citep{yang2025deltanet} blocks and Multi-head Self-attention (MHSA) blocks, especially every 11 DeltaNet blocks followed by one MHSA block. (\romannumeral2) The LLaMA-like macro design~\citep{Touvron2023LLaMA} with RMSNorm~\citep{Zhang2019RMSNorm}, Layer Scale~\citep{iccv2021CaiT}, Rotary Position Embedding (RoPE)~\citep{Su2021RoFormer}, SwiGLU, and FlashAttention to facilitate stable pre-training with long sequences. (\romannumeral3) During pre-training, we apply the packing strategy \citep{Warner2024ModernBERT} to build up a long sequence with several CDS, which compromises the gap between different lengths of the reference sequence and the coding sequences.
With three pre-training tasks in Eq.~\ref{eq:loss_total}, the Life-Code tokenizer and encoder are optimized by the AdamW optimizer for 1M iterations with a batch size of 256 and a basic learning rate of $1\times 10^{-4}$. We adopt 15\% random masking in BERT for Masked DNA Reconstruction and the 3-mer span masking for CDS-to-Amino-Acid Translation. As for KD from a pre-trained Protein LM, we adopted ESM2-650M~\citep{lin2022ESM2} and a protein decoder with an output dimension of 1280. During the warmup periods, the maximum sequence length is 1024, with a linear warmup of the learning rate for 10 iterations. After that, the maximum sequence length is set to 4k with the learning rate adjusted by the Cosine Annealing scheduler.
View Appendix~\ref{app:impl_tokenizer} for more details.

\begin{table*}[t!]
    \centering
    \setlength{\tabcolsep}{0.8mm}
\resizebox{1.0\linewidth}{!}{
    \begin{tabular}{l|cccccccccb}
    \toprule
    Method                           & HyenaDNA & DNABERT & DNABERT2 & GENA-LM & NT-500M    & Caduceus-16 & VQDNA & MxDNA      & ConvNova & \textbf{Life-Code} \\
    \# Params (M)                    & 6.6      & 86      & 117      & 113     & 498        & 7.9         & 93    & 100        & 1.7      & 500                \\
    Model Type                       & RNN      & BERT    & BERT     & BERT    & BERT       & RNN         & BERT  & BERT       & CNN      & Hybrid             \\ \hline
    Enhancers (3 tasks)              & 80.88    & 80.14   & 82.81    & 83.22   & \underline{84.56} & 79.96   & 82.37 & 82.79      & 80.90    & \textbf{86.00}     \\
    Species Classification (2 tasks) & 93.61    & 94.74   & 95.49    & 95.11   & \underline{96.64} & 94.65   & 95.79 & 96.46      & 95.50    & \textbf{97.20}     \\
    Regulatory Elements (3 tasks)    & 88.89    & 83.42   & 86.33    & 87.89   & 89.05      & 85.97       & 87.62 & \underline{90.57} & 87.30    & \textbf{91.30}     \\ \hline
    \bf{Average (8 tasks)}           & 87.07    & 85.02   & 87.30    & 87.94   & \underline{89.26} & 85.89    & 87.69 & 89.12      & 86.95    & \textbf{90.79}     \\
    \bottomrule
    \end{tabular}
    }
    \vspace{-0.5em}
    \caption{\textbf{Comparison on Genomic Benchmarks}.  
    Top-1 accuracy (\%) averaged over several similar tasks is reported for popular DNA foundation models with SFT evaluation.  
    The best and the second best results are marked as the \textbf{bold} and \underline{underlined} types.
    }
    \label{tab:genomic_benchmark}
\end{table*}

% \begin{table*}[t!]
%     \centering
%     \vspace{-0.5em}
%     \caption{\textbf{Comparison on Genomic Benchmarks}. Top-1 accuracy (\%) averaged over several similar tasks is reported for popular DNA foundation models with SFT evaluation, where the best and the second best results are marked as the \textbf{bold} and \underline{underlined} types.
%     }
%     \vspace{1pt}
%     \setlength{\tabcolsep}{0.8mm}
% \resizebox{1.0\linewidth}{!}{
%     \begin{tabular}{l|cccccccccb}
%     \toprule
% % 
% Method                           & HyenaDNA & DNABERT & DNABERT2 & GENA-LM & NT-500M    & Caduceus-16 & VQDNA & MxDNA      & ConvNova & \textbf{Life-Code} \\
% \# Params (M)                    & 6.6      & 86      & 117      & 113     & 498        & 7.9         & 93    & 100        & 1.7      & 500                \\
% Model Type                       & RNN      & BERT    & BERT     & BERT    & BERT       & RNN         & BERT  & BERT       & CNN      & Hybrid             \\ \hline
% Enhancers (3 tasks)              & 80.88    & 80.14   & 82.81    & 83.22   & \ul{84.56} & 79.96       & 82.37 & 82.79      & 80.90    & \textbf{86.00}     \\
% Species Classification (2 tasks) & 93.61    & 94.74   & 95.49    & 95.11   & \ul{96.64} & 94.65       & 95.79 & 96.46      & 95.50    & \textbf{97.20}     \\
% Regulatory Elements (3 tasks)    & 88.89    & 83.42   & 86.33    & 87.89   & 89.05      & 85.97       & 87.62 & \ul{90.57} & 87.30    & \textbf{91.30}     \\
% % 
%     \bottomrule
%     \end{tabular}
%     }
%     \label{tab:genomic_benchmark}
%     \vspace{-0.5em}
% \end{table*}

\begin{table*}[t!]
    \vspace{-0.5em}
    \centering
    \setlength{\tabcolsep}{0.4mm}
\resizebox{1.0\linewidth}{!}{
    \begin{tabular}{l|ccccccccb}
    \toprule
    Method                                 & HyenaDNA & DNABERT & NT-multi & DNABERT2 & Caduceus-PS & VQDNA & MxDNA & ConvNova & \textbf{Life-Code} \\
    \# Params (M)                          & 6.6      & 86      & 2537           & 117      & 1.9         & 93    & 100   & 1.7      & 500                \\ \hline
    Epigenetic Marks Prediction (10)       & 58.94    & 49.08   & 58.06          & 55.98    & 58.39       & 57.95 & 67.29 & \textbf{68.91} & \underline{67.85} \\
    Human TF Detection (3)                 & $-$      & 64.17   & 63.34          & 70.11    & $-$         & \underline{70.56} & $-$ & $-$ & \textbf{71.34} \\
    Mouse TF Detection (3)                 & $-$      & 56.43   & 67.02          & 67.99    & $-$         & \underline{69.80} & $-$ & $-$ & \textbf{71.76} \\
    Core Promoter Detection (3)            & $-$      & 71.81   & 71.63          & 70.53    & $-$         & \underline{73.37} & $-$ & $-$ & \textbf{74.53} \\
    Promoter Detection (3)                 & $-$      & 81.69   & \underline{88.15} & 84.21 & $-$ & 86.58 & $-$ & $-$ & \textbf{89.28} \\
    Splice Site Reconstructed (1)          & $-$      & 84.07   & 89.35          & 84.99    & $-$         & \underline{89.53} & $-$ & $-$ & \textbf{90.24} \\
    Virus Covid Classification (1)         & $-$      & 55.50   & 73.04          & 71.02    & $-$         & \underline{74.32} & $-$ & $-$ & \textbf{74.88} \\ \hline
    \textbf{Average (24 tasks)}            & 58.94    & 60.53   & 67.23          & 66.43    & 58.39       & 68.51 & 67.29 & \underline{68.91} & \textbf{73.51} \\
    \bottomrule
    \end{tabular}
    }
    \vspace{-0.5em}
    \caption{\textbf{Comparison on GUE Benchmark}. Matthews Correlation Coefficient (MCC) (\%) averaged across several sub-tasks is reported with SFT evaluation.  
    The best and the second best results are marked as the \textbf{bold} and \underline{underlined} types.}
    \label{tab:gue}
    \vspace{-0.5em}
\end{table*}

\subsection{Comparison on Genomic Benchmarks}
\label{sec:dna_comparison}
As DNA sequences are the foundation of the central dogma, we primarily compare with the latest DNA foundation models with the SFT evaluation protocol proposed by \citep{nips2024hyenadna, iclr2024dnabert2}, following benchmark evaluations for fair comparison.
We consider three types of DNA model architectures: (a) classical BERT encoders, including DNABERT~\citep{BioInfo2021dnabert}, DNABERT2~\citep{iclr2024dnabert2}, GENA-LM~\citep{fishman2023GenaLM}, Nucleotide Transformer (NT)~\citep{NM2023NucleotideTrans}, with different tokenizers like VQDNA~\citep{icml2024vqdna} and MxDNA~\citep{nips2024MXDNA}; (b) RNN-based linear attentions including HyenaDNA~\citep{nips2024hyenadna} and Caduceus~\citep{icml2024caduceus}; (c) CNN model ConvNova~\citep{iclr2025convnova}.

\vspace{-0.25em}
\paragraph{Genomic Benchmarks.}
We first compare Life-Code with 8 evaluation tasks in the Genomic Benchmark~\citep{BMC2023genomicbenchmark}, which we adopted results from GenBench implementations \citep{liu2024genbench}, and all compared DNA models are fully SFT and reported the top-1 accuracy. Table~\ref{tab:genomic_benchmark} demonstrates that Life-Code surpasses previous models since its hybrid designs combine the advantages.

\begin{table}[t!]
    \centering
    % \setlength{\tabcolsep}{1.0mm}
    % \vspace{-0.5em}
    % \setlength{\tabcolsep}{1.1mm}
\resizebox{0.9\linewidth}{!}{
    \begin{tabular}{l|cccb}
    \toprule
Method    & GenSLM & NT-2500M & EVO  & \bf{Life-Code} \\
\# Params & 2.5B   & 2.5B     & 7B   & 500M           \\ \hline
Bacteria  & 24.7   & 9.4      & \ul{45.3}  & \bf{46.4}      \\
Human     & 6.9    & 4.7      & \ul{11.1}  & \bf{23.8}      \\
    \bottomrule
    \end{tabular}
    }
    \vspace{-0.5em}
    \caption{\textbf{Comparison on Zero-shot Protein Fitness Prediction}. SRCC (\%) is reported on DMS datasets.
    }
    \label{tab:protein_dms}
    \vspace{-0.5em}
\end{table}

\vspace{-0.25em}
\paragraph{GUE Benchmarks.}
We then evaluate Life-Code with 24 long-range genomic tasks on the GUE benchmark~\citep{iclr2024dnabert2}, as shown in Table~\ref{tab:gue}, where 7 widely adopted practical genomic tasks are conducted, including Epigenetic Mark Prediction (EMP) for Yeast, Transcription Factor Prediction on Mouse and Human genome, Promoter Detection, Core Promoter Detection, Splice Site Prediction, and Covid Variants Classification. All results are adopted from DNABERT2 or each paper with their official SFT setups.
% and the Matthews Correlation Coefficient (MCC). 
% Compared to popular DNA models, our Life-Code shows superiority in gene annotation tasks, which are important regulation processes in the central dogma, while achieving competitive results in Epigenetic tasks.

\vspace{-0.25em}
\subsection{Comparison on RNA Splicing Site Prediction}
\label{sec:rna_comparison}
Since precursor messenger RNAs (pre-mRNAs) splicing is a crucial process in eukaryotic gene expression, we then evaluate Life-Code on Splicing Site Prediction tasks as binary classification tasks. We consider two typical datasets with the SFT evaluation protocol.
As for the SpliceAI dataset~\citep{JAGANATHAN2019SpliceAI}, we evaluate DNA models with top-1 AUC scores and find that Life-Code consistently outperforms popular SpliceAI and DNA models, indicating its great ability to model the central dogma. As for Spliceator dataset~\citep{BMC2021spliceator}, we then consider RNA splicing methods (Spliceator, DDSP~\citep{naito2019DDSP}, RINALMo~\citep{Penic2024RiNALMo}) and RNA language model, RNA-FM~\citep{shen2024rnafm}, with RNA data of multi-species. As shown in Table~\ref{tab:splicing}, our Life-Code achieves the best F1 scores.

\begin{table}[t!]
    \centering
\resizebox{1.0\linewidth}{!}{
    \begin{tabular}{l|ccccb}
    \toprule
Dataset    & SpliceAI   & DNABERT2 & NT     & Caduceus      & \textbf{Life-Code} \\
SpliceAI   & 63.2       & 67.1     & 63.9   & \ul{69.1}     & \textbf{70.4}      \\
\toprule
Dataset    & Spliceator & DDSP     & RNA-FM & RINALMo       & \textbf{Life-Code} \\
Spliceator & 91.0       & 91.9     & 92.1   & \ul{94.8} & \bf{95.0}          \\
    \bottomrule
    \end{tabular}
    }
    \vspace{-0.5em}
    \caption{\textbf{Splicing Prediction} with two splicing datasets from SpliceAI and Spliceator. The top-1 AUC score is reported. 
    }
    \label{tab:splicing}
    \vspace{-0.5em}
\end{table}

\vspace{-0.25em}
\subsection{Comparison on Protein Fitness Prediction}
\label{sec:protein_comparison}
Following EVO \citep{nguyen2024evo}, we evaluate foundation models in a strict \emph{zero-shot} setting on \emph{Protein Fitness Prediction}. Deep Mutational Scanning (DMS) datasets \citep{icml2022Tranception,nguyen2024evo} introduce exhaustive mutations into protein coding sequences (CDS) and report fitness scores that reflect functional efficacy. For both bacterial and human proteins, we compute Spearman’s rank correlation (SRCC) between model scores and experimental measurements (Table~\ref{tab:protein_dms}). The protein-specialist ESM-2 sets the upper bound, yet \textbf{Life-Code surpasses all DNA-based baselines} including GenSLM \citep{ijhpca2023genslms} and the 7-B-parameter EVO model—despite using fewer parameters. We attribute this advantage to Life-Code’s unified pre-training on DNA, RNA, and amino acids, which captures translation constraints and folding cues along the central-dogma pathway, enabling strong generalization from nucleotide space to protein fitness without fine-tuning.

\subsection{Comparison on Multi-omics Tasks}
\label{sec:multiomics_comparison}
We conduct further comparisons with proposed biological foundation models (EVO~\citep{nguyen2024evo} and LucaOne~\citep{he2024lucaone}) with multi-omics tasks. Following LucaOne, we first evaluate the non-coding RNA and protein interactions (ncRPI) task using SFT evaluation, which aims to predict the interaction strengths between non-coding RNAs (ncRNAs) and proteins. As shown in Table~\ref{tab:multi_omics}, Life-Code can accept cDNA and Amino Acid sequences in a unified input format and achieve a competitive result, comparable to the supervised LucaOne, which requires separate modules for each input. Then, we evaluate the modeling of transcription and translation processes using binary classification tasks between cDNA and protein sequences, where Life-Code achieves the best accuracy compared to existing models. We can see that the modalities enhance each other.

\begin{table}[t!]
    \centering
\resizebox{1.0\linewidth}{!}{
    \begin{tabular}{l|cc}
    \toprule
Method                & ncRNA-Protein & Central Dogma \\ \hline
DNABERT2+ESM2         & 93.32         & 73.31         \\
EVO                   & $-$           & 75.45         \\
\grow{LucaOne}        & 93.80         & 84.84         \\
\brow{\bf{Life-Code}} & \bf{94.12}    & \bf{86.51}    \\
    \bottomrule
    \end{tabular}
    }
    \vspace{-0.5em}
    \caption{\textbf{Comparison on Multi-omics Tasks}. Top-1 accuracy (\%) is reported for ncRNA-Protein Interaction with SFT and few-shot central dogma evaluation. 
    }
    \label{tab:multi_omics}
    \vspace{-0.25em}
\end{table}

\begin{table}[t!]
    \centering
\resizebox{1.0\linewidth}{!}{
    \begin{tabular}{l|c|cccccc}
    \toprule
    Method       & \# Params   & 1k          & 20k       & 32k       & 100k          & 250k          \\ \hline
    HyenaDNA     & 6.6M        & 61.1        & 87.4      & 93.4      & 96.4          & 97.9          \\
    DNABERT2     & 117M        & 61.0        & 86.8      & \bf{99.3} & \gray{OOM}    & \gray{OOM}    \\
\brow{\bf{Life-Code}} & 500M   & \bf{62.8}   & \bf{88.1} & \bf{99.3} & \textbf{99.4} & \textbf{99.5} \\
    \bottomrule
    \end{tabular}
    }
    \vspace{-0.5em}
    \caption{\textbf{Comparison on Species classification}. Top-1 accuracy (\%) for 5-way classification with scaling up the sequence length, where OOM denotes out-of-memory.
    }
    \label{tab:cls_species}
    \vspace{-0.5em}
\end{table}

\begin{figure}[b!]
  \vspace{-1em}
  \begin{center}
    \includegraphics[width=0.9\linewidth]{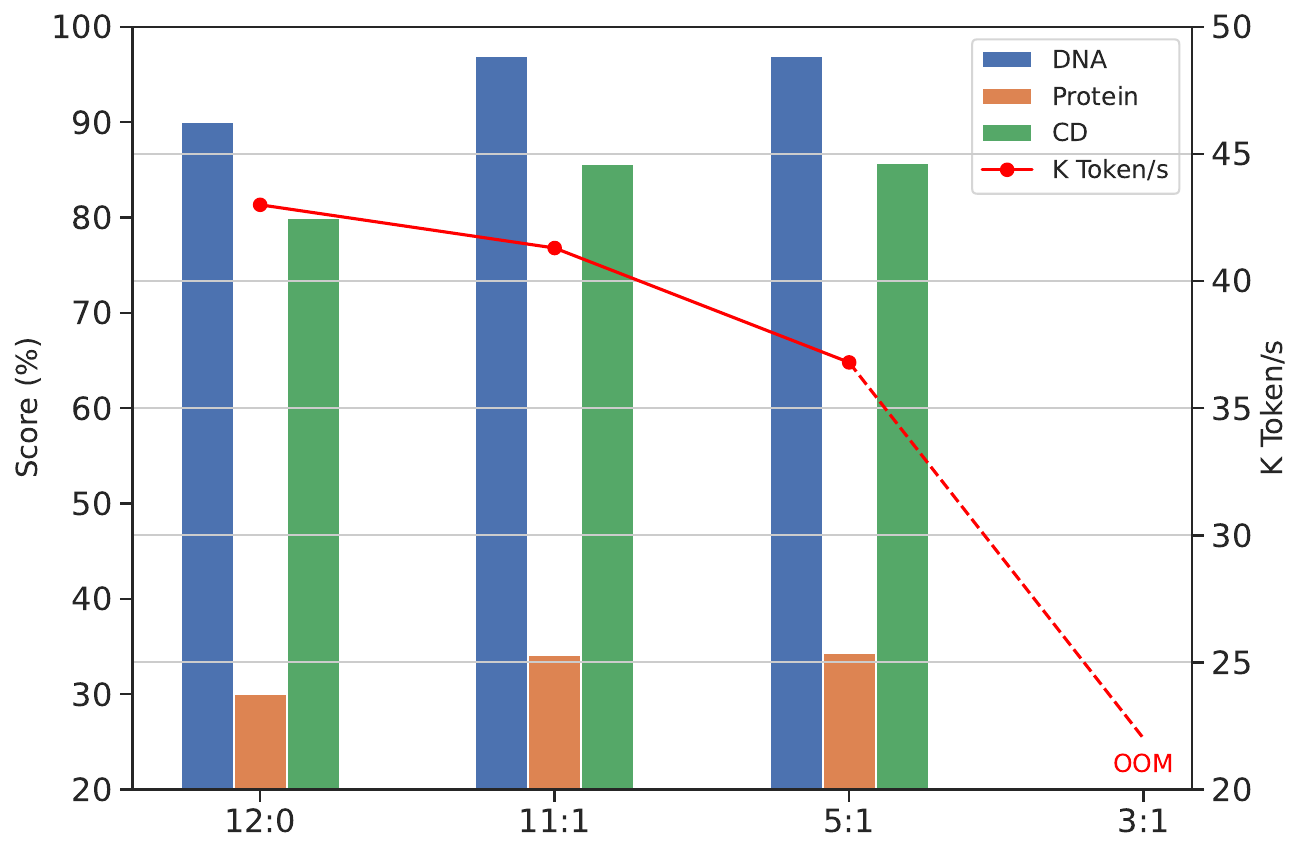}
  \end{center}
  \vspace{-01.5em}
  \caption{Model accuracy vs.\ throughput in K tokens/s for different hybrid layer ratios. A shift from 11:1 to 5:1 slightly slows throughput with minimal accuracy loss.}
  \label{fig:ratio}
\end{figure}

\subsection{Empirical Analysis of Hybrid Design}
\label{sec:exp_empirical}
Figure~\ref{fig:ratio} shows that a pure RNN stack (12{:}0) is fastest (\,$\sim$45 K token/s) but sacrifices 2–3 pp of accuracy.  
Injecting just one MHSA layer (11{:}1) restores nearly all lost performance—DNA 95 to 97 \%, Protein 32 to 34 \%, CD 83 to 86 \%—while reducing throughput by only $\sim$9 \%.  
A 5{:}1 mix boosts accuracy further with a modest extra cost, whereas 3{:}1 exhausts memory.  
{Thus, adding a small fraction of MHSA markedly improves downstream efficiency.}

% \textit{e.g.}, splicing site prediction in Table~\ref{tab:splicing}.
\begin{table}
    \centering
    \setlength{\tabcolsep}{0.8mm}
\resizebox{1.0\linewidth}{!}{
    \begin{tabular}{lcc|ccc}
    \toprule
Tokenizer           & Encoder        & PT Tasks              & DNA       & Protein   & CD        \\ \hline
Linear (1-mer)      & \xmarkg        & MLM                   & 58.3      & $-$       & 28.7      \\
\bcell{GDN (codon)} & \xmarkg        & \bcell{MLM+Trans.}    & 84.5      & 6.4       & 63.2      \\
\bcell{GDN (codon)} & GDN            & MLM+Trans.            & 95.0      & 13.8      & 73.9      \\
\bcell{GDN (codon)} & \bcell{Hybrid} & MLM+Trans.            & \bf{97.2} & 15.4      & 75.7      \\
\brow{GDN (codon)}  & \bcell{Hybrid} & MLM+Trans.+KD         & 97.0      & \bf{34.1} & \bf{85.6} \\
    \bottomrule
    \end{tabular}
    }
    \vspace{-0.5em}
    \caption{\textbf{Ablation Study} of Life-Code and pre-training tasks with DNA, protein, and central dogma (CD) tasks. Note that the blue background denotes the selected setups.
    }
    \label{tab:ablation}
    \vspace{-0.5em}
\end{table}

%%% ablation中DNA task现在结果为species classification，下次改成真正的Covid Virus Classification

\subsection{Ablation Study}
We conduct ablation on the pre-training tasks and the network designs for the tokenizer and encoder, as shown in Table~\ref{tab:ablation}. We consider the top-1 accuracy (\%) Covid Variants Classification in the GUE benchmark for the DNA task, the averaged SRCC (\%) of bacterial and human protein fitness with DMS used in EVO as the protein task, and the top-1 accuracy (\%) of the central dogma evaluation task proposed by LucaOne. Firstly, we verify that the codon tokenizer (instead of 1-mer) could learn more global contextual information by the RNN module and Conv1D (with a kernel size of 3) pre-trained by the MLM task and the cDNA-AA translation task (Trans.). Then, we verified the effectiveness of the hybrid encoder and found that knowledge distillation from ESM2-650M could benefit the protein-related tasks.

\section{Related Works}
\paragraph{DNA Foundation Models.}
% Recent years have seen rapid progress in \textit{DNA} foundation models that leverage Transformer-like architectures. Early attempts, such as DNABERT~\cite{ji2021dnabert}, treated genomic sequences akin to linguistic tokens, enabling pre-trained contextual representations. Follow-up work DNABERT-2~\cite{zhou2023dnabert2} expanded this paradigm by introducing more efficient training protocols and supporting \textit{multi-species} data. In parallel, Nucleotide Transformer \cite{NM2023NucleotideTrans} demonstrated the feasibility of scaling up Transformer architectures for human genomics. Beyond Transformers, kernel-based or hierarchical approaches emerged—\textit{e.g.}, HyenaDNA~\cite{nguyen2024hyenadna}, which reduces the quadratic complexity of attention for extremely long sequences. Meanwhile, alternative designs like Caduceus~\cite{schiff2024caduceus} incorporate selective structural priors for long-range DNA modeling, and MxDNA~\cite{nips2024MXDNA} explores adaptive tokenization schemes that automatically discover suitable patterns for genomic data
Recent DNA foundation models have evolved swiftly from DNABERT \cite{ji2021dnabert}, which first treated genomic 6-mers as linguistic tokens, to DNABERT-2 \cite{zhou2023dnabert2} and the large-scale Nucleotide Transformer \cite{NM2023NucleotideTrans}. To tame quadratic attention on longer sequences, HyenaDNA \cite{nguyen2024hyenadna} replaces full self-attention with efficient hierarchical kernels. Other variants embed biological priors—Caduceus \cite{schiff2024caduceus} for long-range structure and MxDNA \cite{nips2024MXDNA} for adaptive tokenizer.

\vspace{-0.5em}
\paragraph{Protein and RNA Advances.}
AlphaFold \cite{jumper2021alphafold,jumper2021alphafold2} ignited the protein-LM wave, inspiring models like ESM-2 \cite{lin2022ESM2}. RNA efforts followed, e.g., RNA-FM \cite{shen2024rnafm} for 3D structure prediction. While these biomolecule-specific architectures excel individually, recent work now fuses DNA, codon, and protein signals \cite{prakash2024bridging,bioinfo2024genomic,BioInfo2023NTA,nmil2024CaLM}, delivering notable downstream gains \cite{Ren2024COMET} and toward unified models.

% On the \textit{protein} side, AlphaFold series~\cite{jumper2021alphafold,jumper2021alphafold2} revolutionized structure prediction, catalyzing a surge in protein-based language models such as ESM-2~\cite{lin2022ESM2}, which refines large-scale protein embeddings. Beyond proteins, RNA modeling also gained traction; for instance, \cite{shen2024rnafm} employed a language-model-based technique for accurate 3D structure prediction. These efforts underscore the trend toward specialized architectures for each biomolecule yet also highlight the desire for integrative multi-omics solutions.
% More recently, several works~\cite{prakash2024bridging, bioinfo2024genomic} tried to combine or enhance protein models with DNA models and codon information~\cite{BioInfo2023NTA, nmil2024CaLM} and achieve remarkable improvements on downstream tasks~\cite{Ren2024COMET}.

\vspace{-0.5em}
\paragraph{Multi-Omics Modeling.}
Recent studies aim to bridge the gap between \textit{genomic} and \textit{protein} sequences, pushing beyond single-modality tasks. CD-GPT~\cite{zhu2024CDGPT} explicitly connects DNA, RNA, and proteins through the central dogma, while BSM~\cite{xiang2024BSM} highlights the potential for small but effective models covering genes and proteins simultaneously. EVO~\cite{nguyen2024evo} similarly integrates molecular and genome-scale data with a generalized sequence-modeling approach. Closely related is LucaOne~\cite{he2024lucaone}, which advocates a unified nucleic acid and protein language for biological representation.

\section{Conclusion and Limitations}
% We have introduced Life-Code, a method that unifies DNA, RNA, and protein sequences by mapping the latter two back to a nucleotide-based representation. This design offers straightforward \emph{multi-omics} integration and maintains \emph{biological interpretability} via differential encoding of coding versus non-coding regions. To handle the long-range dependencies inherent in genomic data, we employ an efficient \emph{symmetric} convolution-based architecture. Moreover, \emph{knowledge distillation} from large protein models allows our approach to scale without excessive resource demands.
Life-Code unifies DNA, RNA, and protein by remapping all sequences to nucleotides, enabling seamless multi-omics integration while preserving biological meaning (coding vs non-coding). A symmetric convolution backbone captures long-range genomic context, and distilling knowledge from large protein LMs keeps training efficient.

\paragraph{Limitations} Experimental results demonstrate that Life-Code achieves competitive performance in protein structure prediction and phenotype analysis. However, limitations remain: refining the modeling of heterogeneous non-coding regions, incorporating \emph{post-translational modifications}, and addressing the loss of fine-grained information during knowledge distillation. Future work will also focus on integrating additional omics layers (e.g., epigenomics, metabolomics) for a deeper understanding of biological systems. Despite these challenges, we believe \textbf{Life-Code} offers a promising step toward comprehensive \emph{multi-omics} data.

\bibliography{reference}

% \input{./ReproducibilityChecklist/LaTeX/ReproducibilityChecklist.tex}
% \newpage
% \input{ReproducibilityChecklist}

%%%%%%%%%%%%%%%%%%%%%%%%%%%%%%%%%%%%%%%%%%%%%%%%%%%%%%%%%%%%
% \newpage
% \appendix
% \section{Appendix}

\renewcommand\thefigure{A\arabic{figure}}
\renewcommand\thetable{A\arabic{table}}
\setcounter{table}{0}
\setcounter{figure}{0}

\appendix

\newpage
\onecolumn

\section*{Appendix of Life-Code}
% The appendix is structured as follows:
\begin{itemize}[leftmargin=1.25em]
    \item In Appendix~\ref{app:implement}, we provide implementation details of pre-training datasets, network architectures, and training schemes of pre-training and fine-tuning stages with hyper-parameter settings.
    \item In Appendix~\ref{app:additional_res}, we provide detailed descriptions of downstream tasks of biological applications and full comparison results.
\end{itemize}

\begin{figure}[ht]
    % \vspace{-0.5em}
    \centering
    \includegraphics[width=0.65\linewidth]{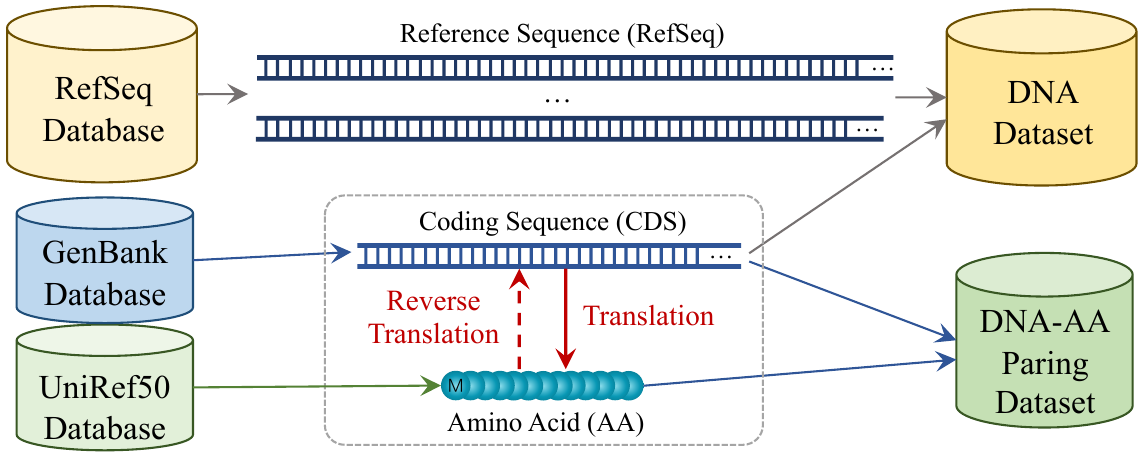}
    % \vspace{-2.0em}
    \caption{\textbf{Data Collection Pipeline}. We collect the reference sequences from the NCBI RefSeq database as the DNA dataset and collect the cDNA of coding sequences with the corresponding amino acids from the GenBank and UniRef50 databases to build our DNA-AA pairing datasets.
    }
    \label{fig:data_collection}
    \vspace{-1.0em}
\end{figure}

% \begin{wraptable}{r}{0.48\linewidth}
% \begin{table}[b!]
\begin{table}[ht]
    \centering
\resizebox{0.525\linewidth}{!}{
\begin{tabular}{l|ccc}
    \toprule
Dateset & Data Type    & Seq Count   & Data Source \\ \hline
DNA     & DNA          & 51,257,875  & NCBI RefSeq \\
        & RNA          & 6,463,852   & NCBI RefSeq \\ \hline
        & cDNA-AA pair & 49,453,295  & GenBank     \\
DNA-AA  & AA-cDNA pair & 17,245,138  & UniRef50    \\
        & mRNA-AA pair & 25,468,310  & GenBank     \\
    \bottomrule
    \end{tabular}
    }
    \caption{
    Configuration of pre-training datasets. As for the DNA dataset, we collect DNA/RNA sequences from the NCBI RefSeq database. As for the DNA-AA pairing dataset, we collect the cDNA (coding sequences) and their corresponding Amino Acids (or using translation) from the GenBank database, and also collect Amino Acids and their corresponding cDNA sequences using reverse translation from the UniRef50 database.
    }
    \label{tab:app_dataset}
    \vspace{-0.5em}
\end{table}
% \end{wraptable}

% % \begin{wraptable}{r}{0.48\linewidth}
% % \begin{table}[b!]
% \begin{table}[ht]
%     % \setlength{\tabcolsep}{1.4mm}
%     \centering
% \resizebox{0.525\linewidth}{!}{
% \begin{tabular}{l|ccc}
%     \toprule
% % 
% Dateset & Data Type    & Seq Count  & Data Source \\ \hline
% DNA     & DNA          & 51,257,875 & NCBI RefSeq \\
%         & RNA          & 6,463,852  & NCBI RefSeq \\ \hline
%         & cDNA-AA pair & 16,786,593 & GenBank     \\
% DNA-AA  & AA-cDNA pair & 17,245,138 & UniRef50    \\
%         & mRNA-AA pair & 3,908,074  & GenBank     \\
% % 
%     \bottomrule
%     \end{tabular}
%     }
%     \caption{
%     Configuration of pre-training datasets. As for the DNA dataset, we collect DNA/RNA sequences from the NCBI RefSeq database. As for the DNA-AA pairing dataset, we collect the cDNA (coding sequences) and their corresponding Amino Acids (or using translation) from the GenBank database, and also collect Amino Acids and their corresponding cDNA sequences using reverse translation from the UniRef50 database.
%     }
%     \label{tab:app_dataset}
%     \vspace{-0.5em}
% \end{table}
% % \end{wraptable}

\section{Implementation Details}
\label{app:implement}
\subsection{Pre-training Dataset}
We collect two datasets for pre-training Life-Code models, \textit{i.e.}, a pure DNA dataset and a DNA-AA pairing dataset, as shown in Figure~\ref{fig:data_collection}.
As for the DNA dataset, we collect the reference sequences (RefSeq) of multiple species to ensure generalization abilities from the database of the National Center for Biotechnology Information (NCBI) at \url{https://www.ncbi.nlm.nih.gov}, following the Multi-species Genomes\footnote{Multi-species Genomes are originally provided in \url{https://huggingface.co/datasets/InstaDeepAI/multi_species_genomes}, which is further extended by DNABERT-2 in \url{https://github.com/MAGICS-LAB/DNABERT_2}} provided by Nucleotide Transformer~\citep{NM2023NucleotideTrans} and DNABERT2~\citep{iclr2024dnabert2}. It includes around 135 species in 7 categories with 35.4B nucleotide bases in total.
As for the DNA-AA pairing dataset, we collect the cDNA of coding sequences (CDS) and their corresponding Amino Acids (AA) in the GenBank database at \url{https://www.ncbi.nlm.nih.gov/genbank}, which aims to model the transcription and translation processes of the central dogma. We also collect some Amino Acids in the UniRef50 database following LucaOne~\citep{he2024lucaone} and obtain their corresponding cDNA by reverse translation with online tools to build cDNA-AA pairs. It contains 92M sequences in total.
We provide detailed information for the used datasets in Table~\ref{tab:app_dataset}.

\subsection{Life-Code Tokenizer}
\label{app:impl_tokenizer}
\paragraph{Vocabulary.}
There are two vocabularies used in Life-Code. The unified vocabulary only uses 4 nucleotides \{A, T/U, C, G\} of nucleic acid with 5 special tokens, including ``[U]"/``[UNK]", ``[PAD]", ``[CLS]", ``[SEP]", and “[MASK]” for unknown nucleotides, padding tokens, the class token, separator tokens, and masking tokens. Meanwhile, the Life-Code can also use the codon vocabulary (\textit{i.e.}, the 3-mer of 4 nucleotides that constructs 64 codon tokens), which could be merged into 20 amino acids of protein (20 uppercase letters excluding ``B", ``J", ``O", ``U", ``X", and ``Z"). It can only be applied when the length of an input sequence is a multiple of 3, \textit{i.e.}, the cDNA of amino acids or matured mRNA (CDS). The pre-trained protein language model employs the amino acid vocabulary of ESM-2 \citep{lin2022ESM2}.

% \begin{wraptable}{r}{0.48\linewidth}
\begin{table}[ht]
    % \vspace{-0.75em}
    % \setlength{\tabcolsep}{1.6mm}
    \centering
\resizebox{0.49\linewidth}{!}{
\begin{tabular}{l|cc}
    \toprule
Configuration       & Tokenizer          & Encoder         \\ \hline
                    % & Tokenizer          & Encoder         \\ \hline
Embedding dim       & 384                & 1024/1280       \\
Block number        & 1                  & 24              \\
GDN blocks          & 1                  & 22              \\
Attention blocks    & 0                  & 2               \\
Attention heads     & 0                  & 16              \\
Parameters          & 8M                 & 305/500M        \\ \hline
Optimizer           & \multicolumn{2}{c}{AdamW}            \\
$(\beta_1,\beta_2)$ & \multicolumn{2}{c}{$(0.9,0.98)$}     \\
Training iterations & 100,000            & 1,000,000       \\
Weight decay        & \multicolumn{2}{c}{$1\times 10^{-2}$}  \\
Base learning rate  & $4\times 10^{-4}$  & $1\times 10^{-4}$ \\
Batch size          & 1024               & 256             \\
LR scheduler        & \multicolumn{2}{c}{Cosine Annealing} \\
Warmup iterations   & 5000               & 10,000          \\
Gradient clipping   & \multicolumn{2}{c}{1.0}              \\
    \bottomrule
    \end{tabular}
    }
    \caption{
    Configuration of the network designs and pre-training settings for Life-Code models. GDN blocks denote the Gated DeltaNet~\cite{yang2024gateddelta} block with linear complexity, while Attention blocks denote the self-attention block with FLASH-Attention implementation~\cite{Dao2022FlashAttention}.
    }
    \label{tab:app_lcode_config}
    \vspace{-0.5em}
\end{table}
% \end{wraptable}

\vspace{-0.5em}
\paragraph{Tokenizer Network.}
As shown in Figure~\ref{fig:tokenizer}, with the nucleotide vocabulary (including 4 nucleic acids and 5 special symbols), the \textit{Life-Code tokenizer} contains the following modules: (a) A linear projection from 9-dim to 384-dim implemented by \texttt{nn.Embedding}; (b) A GDN block (Gated DeltaNet) with 384-dim for global contextual modeling with linear computational complexity; (c) A 1-d Convolution with a kernel size of 3 and stride of 1, followed by an \texttt{UnFold} operation to merge every three nucleotide tokens into 768-dim codon embedding. Similarly, we design the \textit{DNA De-Tokenizer} with the symmetrical network as the Life-Code tokenizer: (a) \texttt{Fold} operation with a 1-d Convolution with a kernel size of 3 to unmerge the codon embedding to 384-dim, (b) A linear projection from 384-dim to 9-dim vocabulary to reconstruct the original DNA sequences. We also design the \textit{Amino Acid Translator} as a two-layer MLP that translates the codon embedding to the corresponding Amino Acid sequences.

\vspace{-0.5em}
\paragraph{Pre-training Settings.}
As shown in Figure~\ref{fig:tokenizer}, we pre-train the Life-Code tokenizer with the DNA de-tokenizer and Amino Acid de-tokenizer by AdamW optimizer for 100,000 iterations (randomly sampled datasets) with a basic learning rate of $2\times 10^{-4}$ and a batch size of 1024, as detailed in Table~\ref{tab:app_lcode_config}. We utilize 8 Nvidia A100-80G GPUs with a per-GPU batch size of 8 and a gradient accumulation time of 16.

\subsection{Life-Code Encoder}
\label{app:impl_encoder}
\paragraph{Encoder Architecture.}
As shown in Table~\ref{tab:app_lcode_config}, the Life-Code encoder has 24 layers in total with the embedding dim of 1024 with the following designs: (1) Mixture of GDN blocks (DeltaNet~\citep{yang2025deltanet}) and multi-head self-attention (MHSA) blocks with the head dim of 64 as a hybrid model, especially every 11 GDN blocks followed by a self-attention block like MiniMax-01~\citep{MiniMax2025MiniMax01}, which could utilize the complementary properties of GDN and MHSA while maintaining efficiency. Notice that the DeltaNet variants~\cite{yang2024gateddelta} introduce short-term depth-wise convolution after the query/key/value projections in the token mixing block, which requires slightly more parameters than the standard Transformer block.
(2) The model macro design employs pre-norm~\citep{acl2019PreNorm} with RMSNorm~\citep{Zhang2019RMSNorm}, Layer Scale~\citep{iccv2021CaiT}, Rotary Position Embedding (RoPE)~\citep{Su2021RoFormer}, SwiGLU~\citep{Touvron2023LLaMA}, and FlashAttention implementations to facilitate training large-scale models stably with long sequences. (3) During pre-training, we apply the packing strategy~\citep{Warner2024ModernBERT} to build up a long sequence with several CDS, which compromises the gap between different lengths of the reference sequence and the coding sequences, as shown in Figure~\ref{fig:data_packing}.

\vspace{-0.5em}
\paragraph{Pre-training Settings.}
As shown in Figure~\ref{fig:lcode_pretraining}, we further pre-train the Life-Code tokenizer and Encoder with three tasks in Eq.~\ref{eq:loss_total} for 1M steps with a batch size of 256 and a basic learning rate of $1\times 10^{-4}$. We adopt 15\% random masking in BERT for Masked DNA Reconstruction and the 3-mer span masking for CDS-to-Amino-Acid Translation. As for Knowledge Distillation from a pre-trained Protein LM, we adopted ESM2-650M (\textit{esm2\_t33\_650M\_UR50D})~\citep{lin2022ESM2} and a protein decoder with the output dimension of 1280. During the warmup periods, the maximum sequence length is 1024, with a linear warmup of the learning rate for 10 iterations. After that, the maximum sequence length is set to 4k with the learning rate adjusted by the Cosine Annealing scheduler (decay to $1\times 10^{-6}$). We utilize 8 Nvidia A100-80G GPUs with a per-GPU batch size of 2 and a gradient accumulation time of 16.

\subsection{Supervised Fine-tuning}
\label{app:impl_SFT}
In most cases, we apply Supervised Fine-tuning (SFT) to transfer pre-trained models to downstream tasks. Following \citep{nips2024hyenadna, iclr2024dnabert2}, adding the decoder head (\textit{e.g.}, an MLP head) to a specific downstream task, the linear attention (RNN) or self-attention blocks in the pre-trained encoder models are frozen, while Low-Rank Adaptation (LoRA) strategy~\citep{Hu2021LoRA} is employed to parameter-efficiently fine-tuning the models by the AdamW optimizer with a batch size of 32. For each task, if the benchmark and models have provided hyper-parameters, we follow the official settings, or we choose the best combinations of the basic learning rate \{$1\times 10^{-5}$, $5\times 10^{-5}$, $1\times 10^{-4}$\}, the weight decay \{0, 0.01\}, the LoRA rank \{4, 8, 16, 24, 48\}, the LoRA alpha \{8, 16, 24, 48, 96\}, and the total fine-tuning epoch \{5, 10\} on the validation set following the GUE benchmark and GenBench~\citep{liu2024genbench}. Note that the maximum input length will be determined for different tasks since the sequence lengths of downstream tasks vary widely. We report the averaged results over three runs with the optimal settings.

\begin{table*}[t]
    \centering
    % \vspace{-0.5em}
    % \vspace{1pt}
    \setlength{\tabcolsep}{0.8mm}
\resizebox{1.0\linewidth}{!}{
    \begin{tabular}{l|cccccccccb}
    \toprule
Method                  & HyenaDNA & DNABERT & DNABERT2  & GENA-LM     & NT-500M     & Caduceus-16 & VQDNA & MxDNA          & ConvNova & \textbf{Life-Code} \\
\# Params (M)           & 6.6      & 86      & 117       & 113         & 498         & 7.9         & 93    & 100            & 1.7      & 350                \\ \hline
Mouse Enhancers         & 79.34    & 80.99   & 81.82     & 82.97       & \ul{85.12}  & 81.63       & 81.06 & 80.57          & 78.40    & \textbf{85.46}     \\
Human Enhancers Cohn    & 72.96    & 70.23   & 75.87     & 75.63       & \ul{76.12}  & 73.76       & 75.63 & 74.67          & 74.30    & \textbf{76.85}     \\
Human Enhancers Ensembl & 90.33    & 89.19   & 90.75     & 91.07       & 92.44       & 84.48       & 90.41 & \ul{93.13}     & 90.00    & \textbf{93.49}     \\ \hline
Coding vs Intergenomic  & 90.97    & 93.64   & 93.58     & 93.24       & \ul{95.76}  & 93.72       & 94.35 & 95.28          & 94.30    & \textbf{96.14}     \\
Human vs Worm           & 96.24    & 95.84   & 97.39     & 96.98       & 97.51       & 95.57       & 97.23 & \ul{97.64}     & 96.70    & \textbf{97.75}     \\ \hline
Human Regulatory        & 93.08    & 88.16   & 87.94     & 88.10       & 93.79       & 87.30       & 90.92 & \textbf{94.11} & 87.30    & \ul{93.93}         \\
Human OCR Ensembl       & 79.14    & 74.96   & 75.82     & 78.98       & 80.42       & \ul{81.76}  & 76.58 & 81.05          & 79.30    & \textbf{82.02}     \\
Human NonTATA Promoters & 94.45    & 87.13   & 95.24     & \ul{96.60}  & 92.95       & 88.85       & 95.37 & 96.56          & 95.30    & \textbf{96.65}     \\
    \bottomrule
    \end{tabular}
    }
    \caption{\textbf{Full Results on Genomic Benchmarks}. Top-1 accuracy (\%) averaged across three trials is reported for the latest DNA foundation models, where the best and the second best results are marked as the \textbf{bold} and \underline{underlined} types.
    }
    \label{tab:app_genomic_benchmark}
    \vspace{-0.5em}
\end{table*}

\begin{table*}[t]
    \centering
    \vspace{-0.5em}
    % \vspace{1pt}
    \setlength{\tabcolsep}{1.0mm}
\resizebox{1.0\linewidth}{!}{
    \begin{tabular}{l|ccccccccb}
    \toprule
Method                     & HyenaDNA & DNABERT & NT-2500M-multi & DNABERT2   & Caduceus-PS & VQDNA          & MxDNA          & ConvNova       & \textbf{Life-Code} \\
\# Params (M)              & 6.6      & 86      & 2537           & 117        & 1.9         & 93             & 100            & 1.7            & 350                \\ \hline
Human TF-0                 & $-$      & 66.84   & 66.64          & \ul{71.99} & $-$         & \textbf{72.48} & $-$            & $-$            & 71.58              \\
Human TF-1                 & $-$      & 70.14   & 70.28          & \ul{76.06} & $-$         & \textbf{76.43} & $-$            & $-$            & 75.92              \\
Human TF-2                 & $-$      & 61.03   & 58.72          & 66.52      & $-$         & \ul{66.85}     & $-$            & $-$            & \textbf{70.63}     \\
Human TF-3                 & $-$      & 51.89   & 51.65          & 58.54      & $-$         & \textbf{58.92} & $-$            & $-$            & \ul{58.74}         \\
Human TF-4                 & $-$      & 70.97   & 69.43          & 77.43      & $-$         & \ul{78.10}     & $-$            & $-$            & \textbf{79.45}     \\ \hline
Mouse TF-0                 & $-$      & 44.42   & \ul{63.31}     & 56.76      & $-$         & 58.34          & $-$            & $-$            & \textbf{64.10}     \\
Mouse TF-1                 & $-$      & 78.94   & 83.76          & 84.77      & $-$         & \ul{85.81}     & $-$            & $-$            & \textbf{86.51}     \\
Mouse TF-2                 & $-$      & 71.44   & 71.52          & 79.32      & $-$         & \ul{80.39}     & $-$            & $-$            & \textbf{80.49}     \\
Mouse TF-3                 & $-$      & 44.89   & 69.44          & 66.47      & $-$         & \ul{69.72}     & $-$            & $-$            & \textbf{71.25}     \\
Mouse TF-4                 & $-$      & 42.48   & 47.07          & 52.66      & $-$         & \ul{54.73}     & $-$            & $-$            & \textbf{55.46}     \\ \hline
Core Promoter (all)        & $-$      & 68.90   & 70.33          & 69.37      & $-$         & \textbf{71.02} & $-$            & $-$            & \ul{70.69}         \\
Core Promoter (no TATA)    & $-$      & 70.47   & \textbf{71.58} & 68.04      & $-$         & 70.58          & $-$            & $-$            & \ul{71.05}         \\
Core Promoter (TATA)       & $-$      & 76.06   & 72.97          & 74.17      & $-$         & \ul{78.50}     & $-$            & $-$            & \textbf{78.78}     \\ \hline
Promoter (all)             & $-$      & 90.48   & \ul{91.01}     & 86.77      & $-$         & 90.75          & $-$            & $-$            & \textbf{91.33}     \\
Promoter (no TATA)         & $-$      & 93.05   & 94.00          & 94.27      & $-$         & \ul{94.48}     & $-$            & $-$            & \textbf{95.03}     \\
Promoter (TATA)            & $-$      & 61.56   & \textbf{79.43} & 71.59      & $-$         & 74.52          & $-$            & $-$            & \ul{78.97}         \\ \hline
Splice Reconstructed       & $-$      & 84.07   & 89.35          & 84.99      & $-$         & \ul{89.53}     & $-$            & $-$            & \textbf{89.76}     \\ \hline
H3                         & 78.14    & 73.10   & 78.77          & 78.27      & 77.90       & 79.21          & \textbf{82.14} & \ul{81.50}     & 81.28              \\
H3K14ac                    & 56.71    & 40.06   & 56.20          & 52.57      & 54.10       & 54.46          & 68.29          & \textbf{70.71} & \ul{68.41}         \\
H3K36me3                   & 59.92    & 47.25   & 61.99          & 56.88      & 60.90       & 61.75          & 65.46          & \textbf{68.31} & \ul{67.25}         \\
H3K4me1                    & 44.52    & 41.44   & 55.30          & 50.52      & 48.80       & 53.28          & 54.97          & \ul{56.60}     & \textbf{57.32}     \\
H3K4me2                    & 42.68    & 32.27   & 36.49          & 31.13      & 38.80       & 34.05          & \ul{55.30}     & \textbf{57.45} & 50.31              \\
H3K4me3                    & 50.41    & 27.81   & 40.34          & 36.27      & 44.00       & 39.10          & \ul{63.82}     & \textbf{67.15} & 53.97              \\
H3K79me3                   & 66.25    & 61.17   & 64.70          & 67.39      & 67.60       & 68.47          & \textbf{73.74} & 72.08          & \ul{72.26}         \\
H3K9ac                     & 58.50    & 51.22   & 56.01          & 55.63      & 60.40       & 56.63          & 63.15          & \textbf{68.10} & \ul{65.45}         \\
H4                         & 78.15    & 79.26   & 81.67          & 80.71      & 78.90       & \ul{81.84}     & 80.89          & 81.12          & \textbf{81.89}     \\
H4ac                       & 54.15    & 37.24   & 49.13          & 50.43      & 52.50       & 50.69          & \ul{65.14}     & \textbf{66.10} & 61.37              \\ \hline
Virus Covid Classification & $-$      & 55.50   & 73.04          & 71.02      & $-$         & \textbf{74.32} & $-$            & $-$            & \ul{73.82}         \\
    \bottomrule
    \end{tabular}
    }
    \caption{\textbf{Full Results on GUE benchmark}. MCC (\%) is reported for Epigenetic Marks Prediction, Human Transcription Factor (TF) Prediction, Mouse Transcription Factor Prediction, Core Promoter Detection, Promoter Detection, Splice Site Reconstructed, and Covid Variants Classification (Virus Covid). The best and the second best results are marked as the \textbf{bold} and \underline{underlined} types.
    }
    \label{tab:app_gue}
    \vspace{-0.5em}
\end{table*}

\section{Downstream Task Settings and Extensive Comparison Results}
\label{app:additional_res}
\subsection{DNA Tasks with Genomics Benchmark}
As proposed by \citep{BMC2023genomicbenchmark}, three groups of basic genomic tasks are collected as binary classification with top-1 accuracy in the Genomics Benchmark. As for the enhancer prediction, three datasets are provided for identifying enhancer regions in the mouse or human genome. As for the species classification, two datasets are selected for identifying sequences as either coding (exonic) or intergenic (non-coding) and classifying sequences as originating from humans or worms (C. elegans). As for the regulatory elements classification, three datasets are used for classifying sequences as regulatory regions based on Ensembl annotations, identifying open chromatin regions, or identifying non-TATA promoter regions in the human genome. We utilize the fully reproduced results of various DNA models in GenBench \citep{liu2024genbench}.

\subsection{DNA Tasks with GUE Benchmark}
As proposed by DNABERT2~\citep{iclr2024dnabert2}, the GUE benchmark contains 24 datasets of 7 practical biological genome analysis tasks for 4 different species using Matthews Correlation Coefficient (MCC) as the evaluation metric. To comprehensively evaluate the genome foundation models in modeling variable-length sequences, tasks with input lengths ranging from 70 to 1000 are selected. The following descriptions of the supported tasks are included in the GUE benchmark, where these resources are attached for illustration.

\textbf{Promoter Detection (Human).}\quad
This task identifies human proximal promoter regions essential for transcription initiation. Accurate detection aids in understanding gene regulation and disease mechanisms. The dataset includes TATA and non-TATA promoters, with sequences -249 to +50 bp around the Transition Start Site (TSS) from Eukaryotic Promoter Database (EPDnew) \citep{dreos2013epdpromoter}. Meanwhile, we construct the non-promoter class with equal-sized randomly selected sequences outside of promoter regions but with TATA motif (TATA non-promoters) or randomly substituted sequences (non-TATA, non-promoters). We also combine the TATA and non-TATA datasets to obtain a combined dataset named \textit{all}.

\textbf{Core Promoter Detection (Human).}\quad
This task is similar to the detection of the proximal promoter with a focus on predicting only the core promoter region, the central region closest to the TSS, and the start codon. A much shorter context window (center -34~+35 bp around TSS) is provided, making this a more challenging task than the prediction of the proximal promoter. 

\textbf{Transcription Factor Binding Site Prediction (Human).}\quad
This task predicts human transcription factor binding sites (TFBS), crucial for gene expression regulation. Data from 690 ENCODE ChIP-seq experiments (161 TF binding profiles in 91 cell lines) \citep{mouse2012encyclopedia} are collected via the UCSC genome browser. TFBS sequences are 101-bp regions around peaks, while non-TFBS sequences match in length and GC content. There are 5 datasets selected from a curated subset of 690, excluding trivial or overly challenging tasks.

\textbf{Splice Site Prediction (Human).}\quad
This task predicts splice donor and acceptor sites, the exact locations in the human genome where alternative splicing occurs. This prediction is crucial to understanding protein diversity and the implications of aberrant splicing in genetic disorders. The dataset \citep{BMC2021spliceator} consists of 400-bp-long sequences extracted from the Ensembl GRCh38 human reference genome. As suggested by \citep{BioInfo2021dnabert}, existing models can achieve almost perfect performance on the original dataset, containing 10,000 splice donors, acceptors, and non-splice site sequences, which is overly optimistic about detecting non-canonical sites in reality. As such, we reconstruct the dataset by iteratively adding adversarial examples (unseen false positive predictions in the hold-out set) in order to make this task more challenging.  
 
\textbf{Transcription Factor Binding Site Prediction (Mouse).}\quad
This task predicts mouse transcription factor binding sites using mouse ENCODE ChIP-seq data (n=78) \citep{mouse2012ChIPseq} from the UCSC genome browser. Negative examples are created by di-nucleotide shuffling. Five datasets are randomly selected from the 78 datasets using the same process as the human TFBS prediction dataset.

\textbf{Epigenetic Marks Prediction (Yeast).}\quad
This task predicts epigenetic marks in yeast, which influence gene expression without altering DNA sequences. Precise prediction of these marks aids in elucidating the role of epigenetics in yeast. We download the 10 datasets from \url{http://www.jaist.ac.jp/~tran/nucleosome/members.htm} and randomly split each dataset into training, validation, and test sets.

\textbf{Covid Variant Prediction (Virus).}\quad
This task aims to predict the variant type of the SARS\_CoV\_2 virus based on 1000-length genome sequences. We download the genomes from the EpiCoV database \citep{khare202SARS_CoV_2} of the Global Initiative on Sharing Avian Influenza Data (GISAID). We consider 9 types of SARS\_CoV\_2 variants, including \textit{Alpha}, \textit{Beta}, \textit{Delta}, \textit{Eta}, \textit{Gamma}, \textit{Iota}, \textit{Kappa}, \textit{Lambda} and \textit{Zeta}.

\begin{table}[htb]
    \centering
    % \vspace{-0.5em}
    % \vspace{1pt}
    % \setlength{\tabcolsep}{0.5mm}
\resizebox{0.75\linewidth}{!}{
    \begin{tabular}{l|cccccb}
    \toprule
Dataset  & SpliceAI   & DNABERT2     & NT   & GENA-LM & Caduceus      & \textbf{Life-Code} \\ \hline
Donor    & 57.4       & 63.5         & 55.7 & 62.9    & \ul{64.2}     & \textbf{64.3}      \\
Acceptor & 69.1       & 70.7         & 72.2 & 73.0    & \ul{74.0}     & \textbf{74.6}      \\
Mean     & 63.2       & 67.1         & 63.9 & 67.9    & \ul{69.1}     & \textbf{70.0}      \\
\toprule
Dataset  & Spliceator & SpliceFinder & DDSP & RNA-FM  & RINALMo       & \textbf{Life-Code} \\ \hline
Human    & 90.0       & 84.5         & 90.6 & 90.7    & \ul{91.3}     & \textbf{91.5}      \\
Fish     & 91.9       & 91.8         & 93.6 & 93.7    & \textbf{97.4} & \ul{97.3}          \\
Fly      & 91.0       & 84.2         & 91.4 & 91.9    & \textbf{95.8} & \ul{94.7}          \\
    \bottomrule
    \end{tabular}
    }
    \caption{\textbf{Full Results of mRNA Splicing Site Prediction}. With two splicing datasets proposed by SpliceAI and Spliceator, the top-1 AUC score or F1 score is reported with DNA models or RNA models, respectively. The best and the second best results are marked as the \textbf{bold} and \underline{underlined} types.
    }
    \label{tab:app_splicing}
    \vspace{-0.5em}
\end{table}

\subsection{mRNA Splicing Tasks}
% \label{app:rna_splicing}
Following \citep{iclr2024dnabert2, shen2024rnafm}, we evaluate pre-mRNA Splicing Site Prediction as the RNA task, which is a crucial process in eukaryotic gene expression. During splicing, introns are removed from precursor messenger RNAs (pre-mRNAs), and exons are joined together to form mature mRNAs. This process is essential for generating functional
mRNAs that could be translated into proteins. Identifying splice sites—the donor sites at the 5' end of introns and the acceptor sites at the 3' end—is vital for accurately predicting gene structure and location.
Concretely, we regard this task as binary classification of RNA splicing site prediction specifically for acceptor sites and consider two splicing datasets in addition to the \textit{Splicefinder} dataset \citep{wang2019splicefinder} used in the GUE benchmark.

\textbf{Spliceator dataset.}\quad
This dataset~\citep{BMC2021spliceator} consists of ``confirmed" error-free splice-site sequences from a diverse set of 148 eukaryotic organisms, including humans. The gold standard dataset GS 1 is adopted, which contains an equal number of positive and negative samples, and the F1 score is used as the evaluation metric. We chose three independent test datasets containing the samples from 3 different species of humans, fish (Danio rerio), and fruit fly (Drosophila melanogaster).

\textbf{SpliceAI dataset.}\quad
This dataset~\citep{JAGANATHAN2019SpliceAI} also constructs a binary classification dataset similar to Spliceator, which utilizes the GTEx (Genotype-Tissue Expression) project for RNA sequencing data from various human tissues and the GENCODE V24lift37 canonical annotation for gene structure information. SpliceAI also references the ClinVar database to evaluate the clinical significance of predicted splicing variants, which contains information on clinically relevant variants and their associations with diseases. This dataset can be regarded as a long-range evaluation and adopts the top-1 AUC-ROC score as the metric.

\begin{table}[htb]
    \centering
    % \vspace{-0.5em}
    % \vspace{1pt}
    % \setlength{\tabcolsep}{0.7mm}
\resizebox{0.67\linewidth}{!}{
    \begin{tabular}{l|ccgb}
    \toprule
Method                     & NT-2500M & EVO-7B & LucaOne       & \textbf{Life-Code} \\ \hline
Promoter (all)             & 91.0     & 88.5   & \textbf{91.6} & 91.3               \\
Virus Covid Classification & 73.0     & 58.7   & \textbf{75.1} & 73.8               \\ \hline
Splice Reconstructed       & 89.4     & 87.5   & 89.1          & \textbf{89.8}      \\ \hline
Bacterial Protein          & 9.4      & 45.3   & \textbf{46.1} & 45.7               \\
Human Protein              & 4.7      & 11.1   & 19.6          & \textbf{22.4}      \\ \hline
Central Dogma              & $-$      & 75.5   & 84.8          & \textbf{85.6}      \\
    \bottomrule
    \end{tabular}
    }
    \caption{\textbf{Full Results of Foundation Models}. As for DNA and RNA tasks, MCC (\%) is reported for Promoter Detection (all), Covid Variants Classification, and Splice Site Reconstructed. SRCC (\%) is reported for Bacterial and Human Protein Fitness Prediction with DMS. Top-1 accuracy (\%) is reported for the Central Dogma evaluation. The best result is marked as the \textbf{bold} type.
    }
    \label{tab:app_overall}
    \vspace{-0.5em}
\end{table}

\subsection{Protein and Multi-omic Tasks}
% \label{app:protein_multiomics}
\paragraph{Zero-shot Protein Fitness Prediction.}
Following EVO~\citep{nguyen2024evo} and protein language models~\citep{lin2023esm}, we employ Deep Mutational Scanning (DMS) studies to evaluate the models' abilities for protein tasks, which introduce many mutations to a protein coding sequence and then experimentally measure the effects of these mutations (as fitness scores) on various definitions of fitness~\citep{icml2022Tranception}. EVO obtained (DMS) datasets with bacterial (prokaryote) and human (eukaryote) proteins from ProteinGYM at \url{https://proteingym.org}.
To adapt this task to nucleotide sequences, EVO proposes to use the wild-type coding sequence and nucleotide mutations reported in the original DMS studies \citep{icml2022Tranception, altae2021widespread}. For generative pre-trained models such as EVO, we rely on likelihood-based scores under the same masking scheme, assessing how well the model anticipates mutations.
The model performances of zero-shot function prediction are measured by the strength of Spearman's Rank Correlation Coefficient (SRCC), which correlates the predicted likelihoods with the experimental fitness measurements.
Full results of protein fitness prediction are shown in Table~\ref{tab:protein_dms} and Table~\ref{tab:app_overall}, in which our Life-Code achieves balancing performances with bacterial and human proteins and surpasses NT-2500M and EVO-7B.

\vspace{-0.5em}
\paragraph{ncRNA-Protein Interactions.}
Following LucaOne~\citep{he2024lucaone}, we consider the multi-omics task of ncRNA-protein interactions (ncRPI), which identifies the interaction strengths between non-coding RNAs (\textit{e.g.}, snRNAs, snoRNAs, miRNAs, and lncRNAs) and proteins. Since experimentally identifying ncRPI) It is typically expensive and time-consuming, but the AI-based ncRPI can be a promising task. LucaOne proposes a binary classification task involving pairs of ncRNA and Amino Acid sequences (20,824 pairs in total) with top-1 accuracy as the metric.

\vspace{-0.5em}
\paragraph{Central Dogma Evaluation.}
To evaluate the modeling of the translation rule in the central dogma, we follow LucaOne~\citep{he2024lucaone} to conduct a binary classification task with top-1 accuracy, which determines whether the DNA sequences and the given proteins are correlated.
LucaOne collects a total of 8,533 accurate DNA-protein pairs from 13 species in the NCBI RefSeq database, with each DNA sequence extending to include an additional 100 nucleotides at both the 5' and 3' contexts, preserving intron sequences within the data. LucaOne generated
double the number of negative samples by implementing substitutions, insertions, and deletions within the DNA sequences or altering amino acids in the protein sequences to ensure the resultant DNA sequences could not be accurately translated into their respective proteins.

% \section*{Impact Statements}
% This paper presents a novel approach, Life-Code, aimed at advancing the field of machine learning and bioinformatics by unifying the modeling of multi-omics data through the guidance of the Central Dogma of molecular biology. By integrating DNA, RNA, and protein data into a single framework, our work has the potential to improve the understanding of complex biological processes and drive advancements in genomics, proteomics, and systems biology.
% \paragraph{Ethical Aspects}
% This work has significant potential applications in fields such as personalized medicine, drug discovery, and disease diagnostics. However, we acknowledge the need to address concerns related to data privacy and the responsible use of sensitive genetic information. We have adhered to ethical guidelines in our experiments, ensuring that no private or identifiable data was used.

\end{document}